\documentclass[10pt,twocolumn,letterpaper]{article}

\usepackage[pagenumbers]{cvpr} % To force page numbers, e.g. for an arXiv version

\usepackage[normalem]{ulem}

\usepackage{threeparttable}
\usepackage{multirow}
\usepackage{adjustbox}
\usepackage{float}
\usepackage{tabularx}
\usepackage{rotating}
\usepackage{booktabs, tabularx}
\usepackage[table]{xcolor}
\usepackage[table]{xcolor}
\usepackage{booktabs}
\usepackage{makecell}
\usepackage{pifont}
\usepackage{arydshln} % for dashed lines
\newcommand{\cmark}{\textcolor{green!60!black}{\ding{51}}} % green check
\newcommand{\xmark}{\textcolor{red}{\ding{55}}} % red cross

\newcommand{\deepfakename}{FakeParts{}}
\newcommand{\datasetname}{FakePartsBench{}}

\usepackage{tcolorbox}
\definecolor{lightgreen}{HTML}{C9EAD5}
\definecolor{lightred}{HTML}{F4C7C3}

\usepackage{comment}
\definecolor{cvprblue}{rgb}{0.21,0.49,0.74}
\usepackage[pagebackref,breaklinks,colorlinks,allcolors=cvprblue]{hyperref}

\def\paperID{} % *** Enter the Paper ID here
\def\confName{CVPR}
\def\confYear{2026}

\title{FakeParts: a New Family of AI-Generated DeepFakes}

\author{
\begin{tabular}{c}
Ziyi Liu$^{1}$\footnotemark[1] \quad
Firas Gabetni$^{3}$\footnotemark[1] \quad
Awais Hussain Sani$^{1}$\footnotemark[1] \quad
Xi Wang$^{2}$\footnotemark[1] \\
Soobash Daiboo$^{1}$ \quad
Ga\"etan Brison$^{1}$ \quad
Gianni Franchi$^{3}$\footnotemark[2] \quad
Vicky Kalogeiton$^{2}$\footnotemark[2] \\
\end{tabular}
\\
\\
$^{1}$Hi!PARIS, Institut Polytechnique de Paris, Palaiseau, France \\
$^{2}$LIX, École Polytechnique, CNRS, Institut Polytechnique de Paris, Palaiseau, France \\
$^{3}$U2IS, ENSTA Paris, Institut Polytechnique de Paris, Palaiseau, France \\
\small \texttt{Project Page:} \href{https://www.lix.polytechnique.fr/vista/projects/2025_fakeparts/}{\textcolor{magenta}{\texttt{https://www.lix.polytechnique.fr/vista/projects/2025\_fakeparts}}}
}

\begin{document}

%%%%%%%%%%
\captionsetup{hypcap=false} % disable for suppress warnings
\twocolumn[{%
\renewcommand\twocolumn[1][]{#1}
\maketitle
\includegraphics[width=1.0\linewidth]{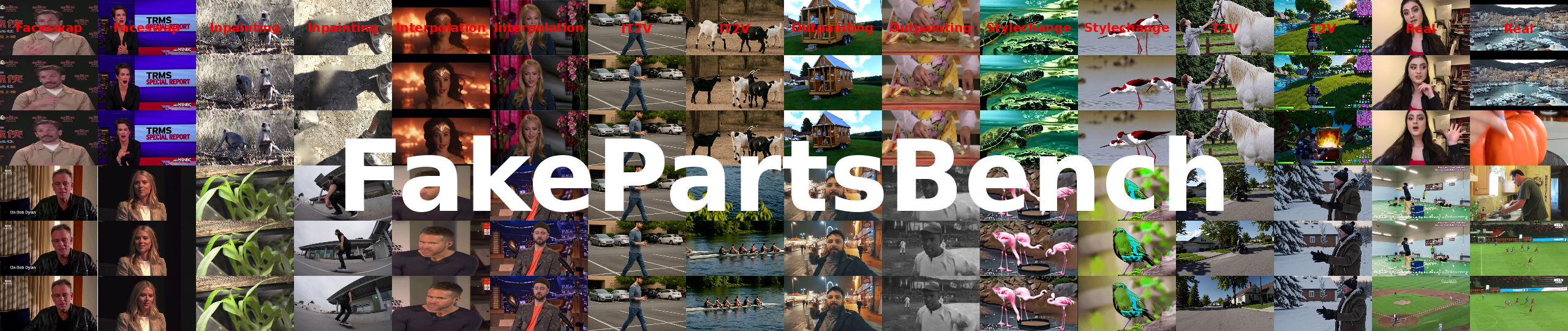}

\captionof{figure}{\textbf{FakePartsBench} is the first dataset specifically designed to include both full deepfakes and FakeParts.
\vspace{2em}}
\label{fig:teaser}
}]
\captionsetup{hypcap=true} % restore afterwards
{
\renewcommand{\thefootnote}{\fnsymbol{footnote}}
\footnotetext[1]{Equal Contribution. \quad$^{\dagger}$ Equal supervision.}
}

\begin{abstract}

We introduce \deepfakename, a new class of deepfakes characterized by subtle, localized manipulations to specific spatial regions or temporal segments of otherwise authentic videos. Unlike fully synthetic content, these partial manipulations—ranging from altered facial expressions to object substitutions and background modifications—blend seamlessly with real elements, making them particularly deceptive and difficult to detect. To address the critical gap in detection, we present \datasetname, the first large-scale benchmark specifically designed to capture the full 
spectrum of partial deepfakes. Comprising over 81K (including 44K FakeParts) videos with pixel- and frame-level manipulation annotations, our dataset enables comprehensive evaluation of detection methods. Our user studies demonstrate that \deepfakename{} reduces human detection accuracy by up to 26\% compared to traditional deepfakes, with similar performance degradation observed in state-of-the-art detection models. This work identifies an urgent vulnerability in current detectors and provides the necessary resources to develop methods robust to partial manipulations.
\end{abstract}

\section{Introduction}

Localized video manipulations present an alarming new frontier in the deepfake landscape. While public attention has focused on fully synthetic videos, this work identifies a more insidious threat: FakeParts, i.e., deepfakes characterized by subtle, localized manipulations affecting only specific spatial regions or temporal segments of otherwise authentic videos.
These partial manipulations, which may alter facial expressions, substitute objects, modify backgrounds, or manipulate individual frames, are particularly dangerous because they blend seamlessly with real content, making them exceptionally difficult to detect. By preserving the majority of the original video, FakeParts leverage the surrounding original, real content to create deceptive content with unprecedented credibility.

The real-world implications of these partial manipulations are concerning. Unlike full-video deepfakes, \deepfakename{} enable targeted reputation attacks where subtle alterations to facial expressions or gestures can change the perceived emotional context of authentic statements. They facilitate sophisticated disinformation campaigns where minimal changes to background elements or object appearances recontextualise events without triggering viewers.
What makes these manipulations timely is their psychological effectiveness: our user study shows that even when explicitly instructed to look for AI-generated\footnote{In this work, we use \textit{AI-Generated} and \textit{deepfakes} interchangeably.} content, human observers fail to identify FakeParts, with detection rates dropping by up to 26\% vs full deepfakes.

This perceptual vulnerability creates a challenge in deepfake detection methods. Yet, current FakePart detection systems are not prepared to address this. This is mostly because no benchmark specifically targets these partial manipulations.
While recent advances have enabled increasingly remarkable video manipulations—with models like Sora~\cite{openai2024sora}, VideoCrafter2~\cite{chen2024videocrafter2}, and ModelScope~\cite{wang2023modelscope} and specialized editing tools like ControlNet-like generation~\cite{guoanimatediff}, video frame interpolation~\cite{wang2024generative,wang2024framerinteractiveframeinterpolation,jain2024video}, camera control~\cite{he2024cameractrl,wang2024motionctrl,wang2024akira} and video inpainter~\cite{zhang2024avid,zi2025cococo}--research on detection methods has lagged behind, primarily due to the lack of representative data.
Specifically, deepfake detection research relies on very large-scale datasets with fully generated content or classic face-swaps~\cite{perov2020deepfacelab,nirkin2019fsgan}, creating a blind spot for more nuanced alterations.

To address this, we introduce \datasetname, the first large-scale benchmark designed to capture the full spectrum of partial deepfakes. It comprises over 81K videos, with 17K real, 20K full deepfakes and 44K FakeParts, which span diverse manipulations, from traditional face-swaps to state-of-the-art generative videos, with emphasis on partial edits, including localized inpainting, style transfer, object substitution, frame-specific edits and temporal interpolation. What distinguishes our dataset is its fine-grained annotation of manipulated regions, providing both pixel- and frame-level precision for evaluating \deepfakename. Drawing from both public and private sources, \datasetname{} offers unprecedented diversity in content, context, and manipulation techniques (Figure~\ref{fig:teaser}).

Our comprehensive evaluation reveals alarming detection gaps for these localized manipulations. Through extensive human studies involving 290 participants, we demonstrate that \deepfakename{} reduces human detection by up to 26\% compared to full-video deepfakes, with certain categories such as interpolation and inpainting manipulations completely undetected. Similarly, state-of-the-art detectors show performance degradation of up to 50\% when confronted with partial manipulations versus fully synthetic videos.
Most concerning is the inverse relationship we discovered between manipulation subtlety and detection difficulty: the smallest edits often produce the most believable deceptions, precisely because they preserve maximum authentic context while changing critical semantic elements.

Our contributions are: (1) We define \deepfakename{} as a novel and increasingly prevalent family of deepfakes distinguished by partial, localized manipulations within otherwise authentic video; (2) We present \datasetname{}, the first benchmark designed to capture partial deepfakes, featuring spatial and temporal manipulations; (3) Out extensive human and detection studies show detection gaps for partial manipulations, set baseline metrics and reveal critical areas for improvement. This work identifies a critical vulnerability in current detectors and provides the foundation for robust defenses against increasingly complex manipulations.

\section{Related Work}

\textbf{Generation and Detection of Deepfake Image}:
Early detectors focused on low-level, hand-crafted inconsistencies such as copy-move, resampling, or sensor noise traces~\cite{Christlein2012,Farid2008,hsu2008video,kobayashi2010detecting,wang2006exposing}.
GANs~\cite{goodfellow2020generative} with large-scale training~\cite{brock2018large} enabled photorealistic faces, leading to detectors that cast the problem as binary classification~\cite{Gragnaniello2021,Marra2018,Yu_2019_ICCV,Zhang2019,tan2023learning}.
However, such detectors often overfit to generators, training data, or image resolution~\cite{Wang_2020_CVPR,Gragnaniello2021}, and such brittleness is amplified for diffusion-generated images~\cite{Bammey2024JoSP,Corvi_2023_CVPR,corvi2023detection}.

Recent detectors can be grouped into:
(i) \textit{Frequency-aware} detectors exploit persistent spectral artifacts~\cite{qian2020thinkingfrequencyfaceforgery, Liu_Li_Zhou_Chen_He_Xue_Zhang_Yu_2021, Wang_Yu_Chen_Hu_Peng_2023, Bammey2024JoSP, ricker446towards, zhong2023patchcraft}.
(i) \textit{Semantics-aware} approaches leverage robust representations from Large Vision-Language models (VLMs), e.g., CLIP~\cite{Cozzolino_2024_CVPR,ojha2023towards,zhang2024common,radford2021learning}.
(iii) \textit{Diffusion-specific} methods probe or invert the generative trajectory~\cite{ma2023exposing,wang2024your}.
Localisation-oriented methods such as FakeLocator~\cite{huang2021fakelocatorrobustlocalizationganbased} predict where manipulations occur in addition to classifying whether an image is fake. 
%further show the value of predicting where manipulations occur, rather than only classifying whether an image is fake.

We posit a unifying framework that places detectors along: a \textit{detection-resolution axis} (pixel/patch~{$\rightarrow$}~image~{$\rightarrow$}~clip) and a \textit{primary-cue axis} (frequency~\cite{huang2021fakelocatorrobustlocalizationganbased, Wang_Yu_Chen_Hu_Peng_2023, qian2020thinkingfrequencyfaceforgery, Liu_Li_Zhou_Chen_He_Xue_Zhang_Yu_2021}, structural~\cite{Li_Bao_Zhang_Yang_Chen_Wen_Guo_2020, Shiohara_Yamasaki_2022, Cozzolino_Rössler_Thies_Nießner_Verdoliva_2021, Xu_Liang_Jia_Yang_Zhang_He_2024}, temporal~\cite{Chen_Akhtar_Haldar_Mian_2025, Güera_Delp_2018, Agarwal_El-Gaaly_Farid_Lim_2020, Masi_Killekar_Mascarenhas_Gurudatt_AbdAlmageed_2020, Sabir_Cheng_Jaiswal_AbdAlmageed_Masi_Natarajan_2019, Shao_Wu_Liu_2022, Tariq_Lee_Woo_2021, Zhang_Lin_Hua_Wang_Zeng_Ge_2022, Liu_2023_WACV, Chu_Xu_You_Zhou_2023, Wang_Bao_Zhou_Wang_Li_2023}, multimodal~\cite{Feng_Chen_Owens_2023, Shao_Wu_Liu_2023, Wang_Wu_Ouyang_Han_Chen_Lim_Jiang_2022}).
This explains the failures of single-scale, single-cue detectors against modern diffusion models and motivates patch-level processing for detecting partial edits.

%%%%%%%%%%%%%%%%%%%%%%%%%%%%%%%%%%%%%%%%%%%%%%%%%%

\begin{table*}[t]
    \centering
    \small
    \begin{adjustbox}{max width=\textwidth}
        \begin{threeparttable}
            
            \begin{tabular}{
                l|    % Dataset
                ccc   % Dur., Real, Fake
                ccrc| % HQ, Meth., Cls-src., FakeParts
                cc|   % FVD\_W, FVD\_FP
                ccc   % VBench: Consist., Flick., Qual.
            }
                \toprule
                \multicolumn{8}{c}{\textbf{Statistics}} &
                \multicolumn{2}{c|}{\textbf{Realism}} &
                \multicolumn{3}{c}{\textbf{VBench Metrics}} \\
                \cmidrule(r){1-8} \cmidrule(r){9-10} \cmidrule(r){11-13}
                \textbf{Dataset} &
                \textbf{Dur.(s)} & \textbf{Real} & \textbf{Fake} &
                \textbf{Hi-Res} & \textbf{\#Meth.} & \textbf{Cls-src.} & \textbf{FakeParts} &
                \textbf{FVD\_W\textsuperscript{}} & \textbf{FVD\_FP\textsuperscript{}} &
                \textbf{Cons.} & \textbf{Flick.} & \textbf{Qual.} \\
                \midrule
                GVD~\cite{ma2025detectingaigeneratedvideoframe} &
                3 & 0 & 11,618 & 5,598 & 11 & 48 & \xmark &
                387.7 & 549.7 & 0.932 & 0.968 & \textbf{0.656} \\

                VidProM~\cite{wangvidprom} &
                3 & 0 & 6,690,00\textsuperscript{1} & 30,000\textsuperscript{1} & 4 & 0 & \xmark &
                290.4 & 428.5 & 0.916    & 0.946 & 0.605    \\

                GenVidBench~\cite{ni2025genvidbench} &
                2 & 13,800 & 55,200 & 0 & 4 & 13,500\textsuperscript{2} & \xmark &
                558.9 & 658.8 & 0.932 & 0.964 & 0.448 \\

                DeMamba~\cite{chen2024demamba} &
                4 & 10,000 & 8,588 & 2,156\textsuperscript{3} & 10 & 2,082 & \xmark &
                382.8 & 459.7 & 0.934 & 0.967 & 0.651 \\

                % \zliu{DigiFakeAV}~\cite{Liu_Wang_Hou_Ren_Wu_He_2025} &
                % 5--10 & 10,000 & 50,000 & 0 & 6 & 3 & \xmark &
                % 301.2 & -- & 0.940 & 0.970 & 0.623 \\
                \midrule
                \textbf{\datasetname{}} &
                5 & 17,252 & 64,252 & $\geq$24,365 & 21 & 12,000 & \cmark &
                \textbf{240.8} & \textbf{211.5} & \textbf{0.940} & \textbf{0.970} & 0.623 \\
                \bottomrule
            \end{tabular}%
            \begin{tablenotes}[flushleft]
                % \item[*] HS refers to high-quality fake data, defined as having a resolution of at least 720p.
                \item[1] VidProM Hi-Res comprises 30,000 videos at 720p from StreamingT2V, Open-Sora 1.2, and CogVideoX-2B. As it lacks a test set, we use these plus 10\% randomly sampled from the training set.
                \item[2] GenVidBench's closed-source data are 560p at 24 fps, generated using Pika in 2022.
                \item[3] DeMamba's Hi-Res includes 700 samples from MorphStudio, 1,400 from Lavie (both from 2023), and 56 videos from Sora.
                % \item[4] DigiFakeAV~\cite{Liu_Wang_Hou_Ren_Wu_He_2025}: 60K talking-head clips (10K real, 50K fake, 5–10 s, $512\times512$ px). Includes RV-RA, FV-RA, and FV-FA settings using six generation models (Sonic, Hallo 1/2, Echomimic, V-Express, CosyVoice 2).
                % \item[*] FVD\_W and FVD\_FP denote FVD computed against WebVid-10M and our collected real dataset, respectively.
            \end{tablenotes}
        \end{threeparttable}
    \end{adjustbox}
    \caption{\textbf{Statistics of test sets of diffusion-based Deepfake Video Datasets.} FVD\_W and FVD\_FP denote FVD computed against WebVid-10M and our collected real dataset, respectively.
    %\zliu{I started running VBench, no idea when it finishes, I'll check tomorrow}}
    }
            \label{tab:landscape-diffusion-based}
\end{table*}

\noindent  \textbf{Image Deepfake Datasets.}
%
%have evolved alongside generative models.
Early datasets focused on GAN-generated faces~\cite{karras2019style}, e.g., ForenSynths~\cite{Wang_2020_CVPR}.
Subsequent corpora include DFDC~\cite{dolhansky2020deepfake}, FaceForensics++~\cite{rossler2019faceforensics++}, WildDeepfake~\cite{Zi_Chang_Chen_Ma_Jiang_2021}, and KoDF~\cite{kwon2021kodf}.
Recent ones target diffusion-generated images at scale, such as GenImage~\cite{Zhu_Chen_Yan_Huang_Lin_Li_Tu_Hu_Hu_Wang_2023}, DiffusionFace~\cite{Chen_Sun_Zhou_Lin_Sun_Cao_Ji_2024}, and Deepfake-Eval-2024~\cite{Chandra_Murtfeldt_Qiu_Karmakar_Lee_Tanumihardja_Farhat_Caffee_Paik_Lee_et_al_2024} 
and face two issues: (1) they assume that edited regions are semantically dominant (e.g., full-face edits); and 
(2) only few provide fine-grained masks for small or blend-in edits.
These motivate the shift to videos, where partial edits are more realistic and harder to detect.

\begin{table}[t]
    \centering
    \resizebox{\linewidth}{!}{
    %\begin{threeparttable}
        %\footnotesize
        \begin{tabular}{l l l l| c c c}
            \toprule
            \multicolumn{4}{c}{\textbf{Metadata}} & \multicolumn{3}{c}{\textbf{\deepfakename{}}} \\
            \cmidrule(r){1-4} \cmidrule(l){5-7}
             & \textbf{Name} & \textbf{Type} & \textbf{Venue} & \textbf{Spatial} & \textbf{Temporal} & \textbf{Style} \\
            \midrule
            \multirow{6}{*}{\rotatebox[origin=c]{90}{GAN}}
            & ForgeryNet~\cite{he2021forgerynet} & Face &  CVPR'21 & \cmark & \xmark & \xmark \\
            & FakeAVCeleb~\cite{khalid2fakeavceleb} & F\&V &  NeurIPS'21 & \cmark & \xmark & \xmark \\
            & SWAN-DF~\cite{korshunov2023vulnerability} & F\&V & IJCB'23 & \cmark & \xmark & \xmark \\
            & DiffSwap~\cite{zhao2023diffswap} & Face & CVPR'23 & \cmark & \xmark & \xmark \\
            & AV-Deepfake1M~\cite{cai2024av} & F\&V & MM'24 & \cmark & \xmark & \xmark \\
            & DF40~\cite{Yan_Yao_Chen_Zhao_Fu_Zhu_Luo_Wang_Ding_Wu_et_al_2024} & Face & NeurIPS'24 & \cmark & \xmark & \xmark \\
            \midrule
            \multirow{5}{*}{\rotatebox[origin=c]{90}{Diffusion}}
            & GVD~\cite{bai2024ai} & Gen & PRCV'24 & \xmark & \xmark & \xmark \\
            & VidProM~\cite{wangvidprom} & Gen & NeurIPS'24 & \xmark & \xmark & \xmark \\
            & GenVidBench~\cite{ni2025genvidbench} & Gen & arXiv'25 & \xmark & \xmark & \xmark \\
            & DeMamba~\cite{chen2024demamba} & Gen & arXiv'25 & \xmark & \xmark & \xmark \\
            % & \zliu{DigiFakeAV}~\cite{Liu_Wang_Hou_Ren_Wu_He_2025} & Face \& Voice & 2025 & arXiv & \xmark & \xmark & \xmark \\
            %\cmidrule(lr){2-8}
            & \textbf{\datasetname{} (Ours)} & Gen & 2025 & \cmark & \cmark & \cmark \\
            \bottomrule
        \end{tabular}
    %\end{threeparttable}
    }
            \caption{\textbf{Landscape of Deepfake Video Datasets.} F\&V corresponds to Face \&Voice and Gen to Generic videos, such as scenes, interior and exterior landscapes, cities, faces, etc.}
        \label{tab:Deepfakereletadworks}
\end{table}

%%%%%%%%%%%%%%%%%%%%%%%%%%%%%%%%%%%%%%%%%%%%%%%%%%

\noindent \textbf{Generation and Detection of Video Deepfakes}:
Early video forgeries primarily involve face swapping, i.e., Deepfakes mainly based on GANs~\cite{nirkin2019fsgan, Shen_2020_CVPR, Tewari_2020_CVPR, Thies_2016_CVPR, tian2018cr, Wu_2018_ECCV, Zhu_2017_ICCV}.
Video detectors initially mirrored image-based strategies: framewise classifiers trained on real/fake clips~\cite{guarnera2020deepfake,li2018exposing,qian2020thinking,rossler2019faceforensics++,Zhao_2021_CVPR}.
Temporal reasoning was later introduced through RNNs for dynamics~\cite{Güera_Delp_2018} or segment localisation~\cite{Masi_Killekar_Mascarenhas_Gurudatt_AbdAlmageed_2020}, and joint appearance-behaviour cues~\cite{Agarwal_El-Gaaly_Farid_Lim_2020}.
With diffusion-based video generators emerging~\cite{blattmann2023stable,openai2024sora,google2025veo2,zhou2024allegro}, newer detectors integrate semantic backbones and VLMs~\cite{song2024learning,ma2025detectingaigeneratedvideoframe}, or spatio-temporal fusion such as optical flow~\cite{bai2024ai}.
Recent methods like DeMamba~\cite{chen2024demamba}, built upon Mamba~\cite{gu2023mamba}, show that long-range inconsistencies can be captured efficiently.

Within the aforementioned two-axis framework, these approaches extend the detection-resolution axis (frame~$\rightarrow$~clip) but remain sensitive to local cues such as lip-speech synchronisation~\cite{zhang2023ummaformer, Zhou_2021_ICCV, haliassos2021lips}.
Consequently, they struggle with \textit{partial, short-span} manipulations, where only a few frames are altered and edits may not be facial.

%%%%%%%%%%%%%%%%%%%%%%%%%%%%%%%%%%%%%%%%%%%%%%%%%%

\noindent \textbf{Video Deepfake Datasets} 
have evolved to large scale, e.g. FF++ video splits~\cite{rossler2019faceforensics++}, DFDC~\cite{dolhansky2020deepfake}, and DeeperForensics-1.0~\cite{Jiang_Li_Wu_Qian_Loy_2020}, which target video face-manipulation under realistic capture and compression. Recent efforts aim at broader and newer forms of AI-generations: AV-Deepfake1M~\cite{cai2024av} scales to the audio-visual setting; 
DF40~\cite{Yan_Yao_Chen_Zhao_Fu_Zhu_Luo_Wang_Ding_Wu_et_al_2024} assesses detectors' generalisation across unseen forgeries and domains;
while the recent~\cite{Liu_Wang_Hou_Ren_Wu_He_2025} targets multimodal deepfake detection for digital humans. In parallel, very recent datasets such as VidProM~\cite{wangvidprom}, GenVidBench~\cite{ni2025genvidbench}, GVD/AIVGDet~\cite{bai2024ai}, and DeMamba~\cite{chen2024demamba} explicitly explore T2V/I2V diffusion models, multimodal generation, and prompt configurations, making them suitable testbeds for fully synthetic videos.

These datasets exploit large-scale or multimodal generation~\cite{ni2025genvidbench,wangvidprom,bai2024ai} with optimized prompts~\cite{wangvidprom}. Yet, they focus solely on fully synthetic videos (via T2V or I2V), omitting partially manipulated content (Table~\ref{tab:Deepfakereletadworks}), and often lack modern high-resolution, high-quality samples (Table~\ref{tab:landscape-diffusion-based}). These limitations not only limit progress towards robust detectors, but also fail to reflect real-world video forgery scenarios, which increasingly involve partial edits and content generated by high-quality proprietary models.

\section{What is missing from Deepfake datasets?}

Video Deepfake methods span a wide spectrum, not only in terms of underlying methodology (e.g., GAN~\cite{karras2019style,nirkin2019fsgan, Shen_2020_CVPR, Tewari_2020_CVPR, Thies_2016_CVPR, tian2018cr, Wu_2018_ECCV, Zhu_2017_ICCV} vs. diffusion~\cite{blattmann2023stable, openai2024sora, google2025veo2, zhou2024allegro}), but also in the nature of input conditions, such as text-to-video (T2V)~\cite{chen2024videocrafter2,openai2024sora,google2025veo2,zhou2024allegro}, image-to-video (I2V)~\cite{blattmann2023stable,guoanimatediff,karras2023dreampose}, or a combination of both (TI2V)~\cite{openai2024sora,google2025veo2}. They also differ in chromatic styling~\cite{wang2023stylediffusion}, spatial manipulation strategies, ranging from face swapping~\cite{perov2020deepfacelab,zhao2023diffswap} to regional inpainting~\cite{reda2022film} and outpainting~\cite{he2024cameractrl,wang2024akira}, as well as in temporal operations like frame interpolation~\cite{wang2024framerinteractiveframeinterpolation}.

To better understand the current landscape of video Deepfake datasets, we summarize key datasets in Table~\ref{tab:Deepfakereletadworks} and observe \textbf{one main limitation}: 

\begin{itemize}[leftmargin=*]
    \item \textbf{Limitation 1: \deepfakename{}}. Modern datasets lack videos with partial manipulations, referred to as FakeParts, which were prevalent during the GAN era but are now mostly absent from diffusion-based video benchmarks.
\end{itemize}

To evaluate recent \textit{diffusion-based} datasets, we benchmark two key aspects: \\
\noindent \textbf{(i) generation quality} through Fréchet Video Distance (FVD)~\cite{ge2024content} computed with respect to two real reference distributions: WebVid-10M~\cite{Bain21} and our included Real data, each sampled 10,000 datapoints; 
and \\
\noindent \textbf{(ii) consistency, temporal flicker and image quality on VBench~\cite{huang2024vbench}}, corresponding to general video performance. Table~\ref{tab:landscape-diffusion-based} summarizes these results along with dataset statistics, including total duration, number of test samples, generation methods (\#Method), high-resolution samples ($>$720p), and closed-source content. Details in the suppl.\ material. From Table~\ref{tab:landscape-diffusion-based}, we observe the second limitation: 

\begin{itemize}[leftmargin=*]
    \item \textbf{Limitation 2: Videos have low perceptual quality.} We observe a clear correlation between low resolution and low perceptual quality. For comparison, VidProM~\cite{wangvidprom}, which includes a higher proportion of high-resolution content, achieves significantly better FVD scores than other existing datasets.
\end{itemize}

\begin{table*}[ht!]
\begin{minipage}[b]{0.72\textwidth}
\centering
\resizebox{\linewidth}{!}{
    \small{
    %\begin{tabular}{ll rr rr r}
     \begin{tabular}{p{2.7cm} c r r >{\bfseries}r l l l}
        \toprule
        Task & \#Meth.\ & Real V. & Fake V. & Total & Resolutions & FPS & Duration \\
        \midrule
        Style Change & \textbf{2}
        & 0 & 5,266 & 5,266 & 512$\times$512; & 8--10 & 2--14 \\
        \footnotesize{\cite{ku2024anyv2v, kara2023raverandomizednoiseshuffling}} &  & &  & & 680$\times$384 & &  \\
        
        Extrapolation & \textbf{1}
        & 120,000 & 3,000 & 3,000 & 1280$\times$704 & 25 & 5 \\
        ~{\footnotesize{\cite{nvidia2025cosmospredict2github, nvidia2025cosmosworldfoundationmodel}}} & & & &  & & \\
        
        Faceswap~{\footnotesize{\cite{deng2018arcface}}} & \textbf{1}
        & 0 & 5,036 & 5,036 & --- & 30 & 11 \\

        Inpainting & \textbf{3}
        & 7,252 & 13,577 & 20,829 & 426$\times$320 to & 5--6 & 4--6 \\
        {\footnotesize{\cite{li2025diffueraserdiffusionmodelvideo, zhou2023propainter, wu2024languagedrivenvideoinpaintingmultimodal}}}
        & &  & & & 1280$\times$720 & &  \\

        Interpolation~{\footnotesize{\cite{wang2024framerinteractiveframeinterpolation}}} & \textbf{1}
        & 0 & 10,000 & 10,000 & 512$\times$320 & 7 & 7 \\

        Outpainting~{\footnotesize{\cite{wang2024akira}}} & \textbf{1}
        & 0 & 7,275 & 7,275 & 576$\times$320 & 8 & 2 \\

        Real & \textbf{1}
        & 10,000 & 0 & 10,000 & 640$\times$360 & 30 & 6 \\

        T2V~{\footnotesize{\cite{yang2025cogvideoxtexttovideodiffusionmodels, kong2025hunyuanvideosystematicframeworklarge, hacohen2024ltxvideorealtimevideolatent, genmo2024mochi}}} & \textbf{9}
        & 0 & 16,119 & 16,119 & 720$\times$480 to & 15--30 & 4--6 \\
        {\footnotesize{\cite{zheng2024opensorademocratizingefficientvideo, wan2025wanopenadvancedlargescale, zhou2024allegro, openai2024sora, google2025veo2}}} & 
        & &  & & 1920$\times$1080 & & \\

        Text-to-Image-to- & \textbf{2}
        & 0 & 3,980 & 3,980 & 720$\times$1280; & 16--24 & 5 \\
      Video  {\footnotesize{\cite{zheng2024opensorademocratizingefficientvideo, wan2025wanopenadvancedlargescale}}}& & & & & 1024$\times$576 & & \\

        \midrule
        \textbf{ALL} & \textbf{21}
        & \textbf{17,252} & \textbf{64,252} & \textbf{81,504}
        & -- & -- & -- \\
        \bottomrule
    \end{tabular}}
    }
    \captionof{table}{
    Statistics and Quality of \datasetname{} Grouped by Task Category.
    }
     % \vic{caption unclear, Ziyi please fix}}
     \label{tab:video_counts}
\end{minipage}
\hfill
\begin{minipage}[b]{0.26\textwidth}
    \centering
    \vspace{-4mm}
    \includegraphics[width=0.95\textwidth]{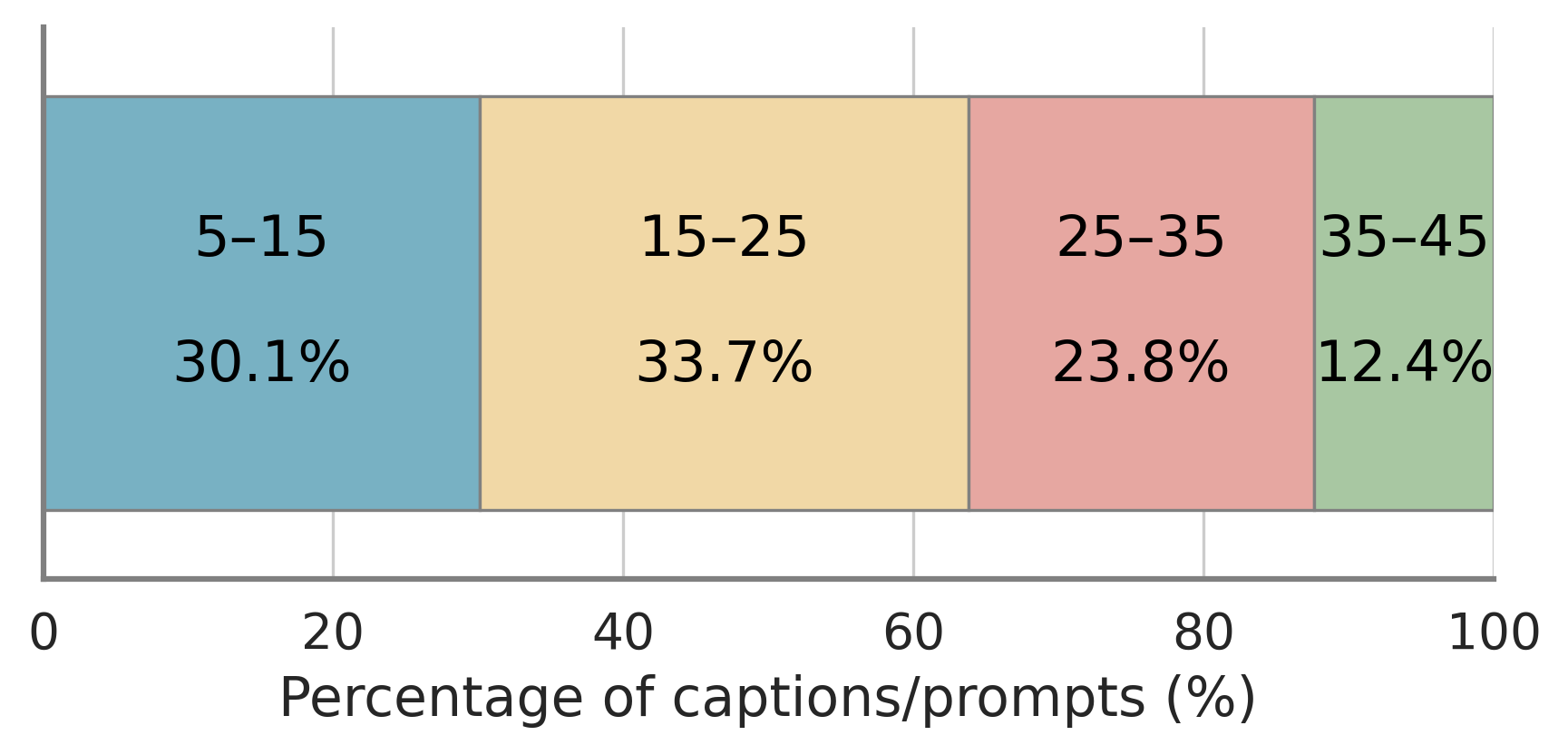}
    \vspace{-1mm}
    \captionof{figure}{Prompt word-count.}
    \label{fig:by_wordcount}
 %\vspace{-1mm} \vspace{1em} 
  % second sub-figure
  %
    % \includegraphics[width=0.9\textwidth]{figures/dataset/distribution_of_topics_dataset.png}
    % \vspace{-1mm}
    % \captionof{figure}{Topic distribution.}
    \includegraphics[width=0.95\linewidth]{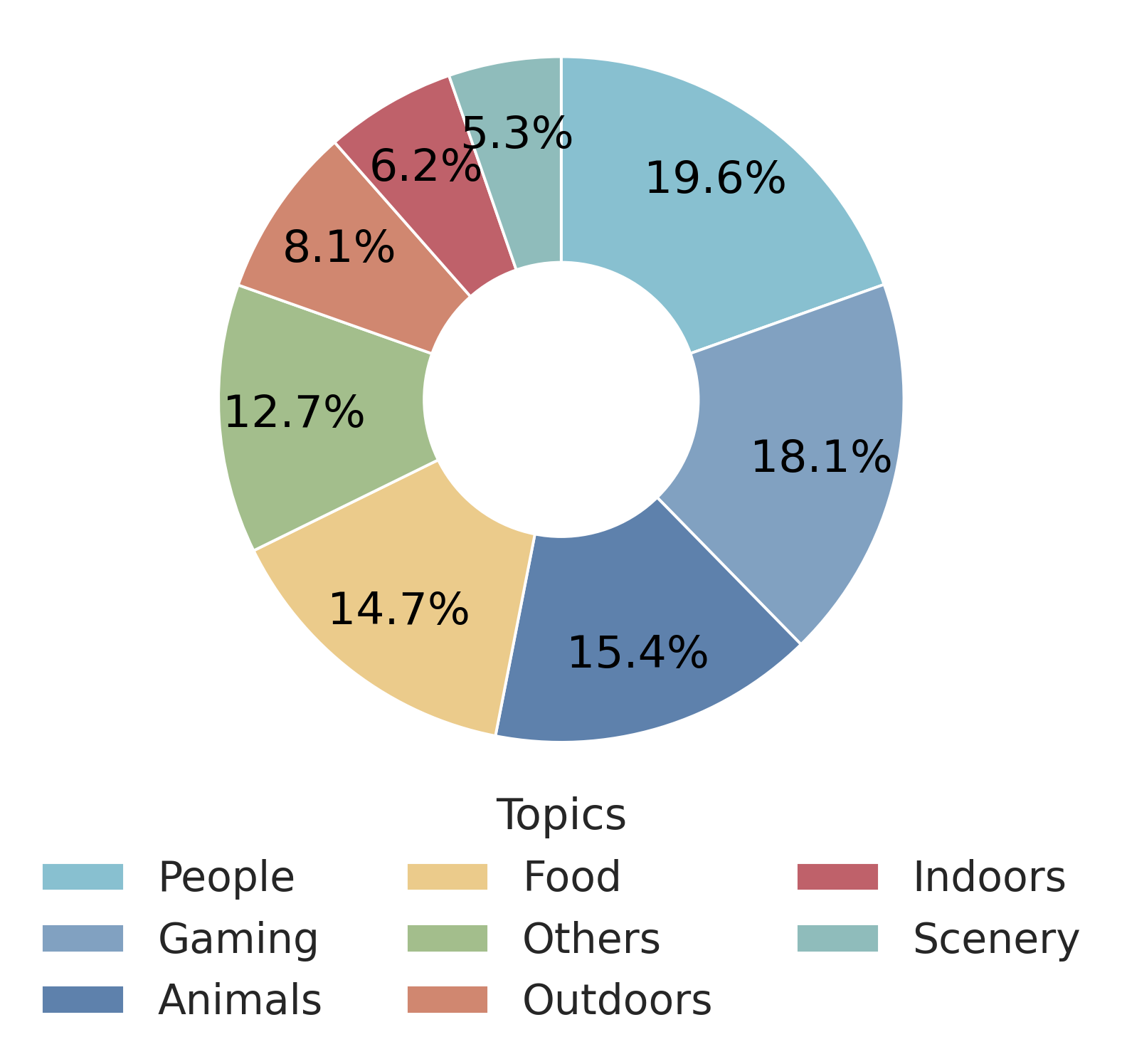}
    \vspace{-1mm}
    \caption{Prompt topic distribution.}
    \label{fig:by_topic}
  
\end{minipage}
  % 
  % \label{fig:combined_distributions}

    % \captionsetup{skip=0cm, width=0.8\textwidth}
    % \captionof{table}{(left) (right) Caption length and topic distribution of generated content in \datasetname{}.}
    % \label{tab:noidea}

\end{table*}
\section{\datasetname: A Benchmark for Video Deepfakes and \deepfakename{}}\label{sec:Benchmarkinfo}

In this work, we introduce \datasetname, a benchmark designed to address the two aforementioned limitations by 
\begin{itemize}[leftmargin=*]
    \item incorporating a diverse and balanced mix of both full and partial manipulations ({\deepfakename}); 
    \item containing videos with high resolution and high quality. For this, we include videos generated by the latest open-source and closed-source models, such as Sora~\cite{openai2024sora}, Veo2~\cite{google2025veo2}, and Allegro~\cite{zhou2024allegro}. This ensures the timeliness and complementarity with existing benchmarks and also better reflects real-world scenarios, where an ever-growing share of Deepfake content online is produced by proprietary systems, often involving partial edits.

\end{itemize}

\noindent {\datasetname} supports research on detecting overt and covert forgeries. By combining large-scale coverage, high-resolution and fine-grained annotations, it enables rigorous evaluation. Next, we give statistics (Section~\ref{sub:statistics}), detail its generation pipeline (Section~\ref{sub:generation}) and define \deepfakename{} as:

\noindent \textbf{(i) Full Deepfakes} include entirely generated or heavily modified content, typically synthesized from a single input modality: e.g., text-to-video (T2V), or text-to-image-to-video (TI2V) (see Section~\ref{paragraph: full deepfake} for details).

\noindent \textbf{(ii) \deepfakename} include fine-grained manipulations that affect only specific aspects of a video. We further divide these into three subtypes (further elaborated in Section~\ref{paragraph: full deepfake}):
    \begin{enumerate}[leftmargin=*]
        \item \textbf{Spatial \deepfakename:} Manipulations on specific regions, e.g., face swapping (FaceSwap), object modification (inpainting), or contextual expansion (outpainting).
        
        \item \textbf{Temporal \deepfakename:} Edits along the temporal axis, e.g., frame interpolation.
        
        \item \textbf{Style \deepfakename:} Changes in visual appearance without altering structural content: e.g., modifying color schemes or applying different visual styles.
    \end{enumerate}

\subsection{\datasetname{} Statistics}
\label{sub:statistics}
Table~\ref{tab:video_counts} reports statistics of \datasetname{}. It contains approximately 81K video clips, split into 17K real and \textbf{64K fake} clips; hence, fake clips constitute more than 80\% of the dataset, reflecting the broad coverage of edited and generative video categories represented.

\datasetname{} spans \textbf{9 tasks} from 21 distinct methods: \emph{Style Change}, \emph{Extrapolation}, \emph{Faceswap}, \emph{Inpainting}, \emph{Interpolation}, \emph{Outpainting}, \emph{Real}, \emph{T2V} (text-to-video), and \emph{TI2V} (text-with-image-to-video). These tasks cover both traditional manipulation pipelines (e.g., Faceswap and Inpainting), modern generative video synthesis settings (T2V/TI2V) and novel partial-fake generation, including object colour editing and extrapolations. Resolutions range from 426$\times$320 to 1920$\times$1080, frame rates from 50 to 30 FPS, and clip durations from 2 to 14 secs.

Importantly, \datasetname{} contains $>$\textbf{24K} high-resolution videos ($\geq$720p), accounting for 30\% of the dataset.

\begin{figure*}[ht]
  \centering
  \includegraphics[width=0.95\linewidth]{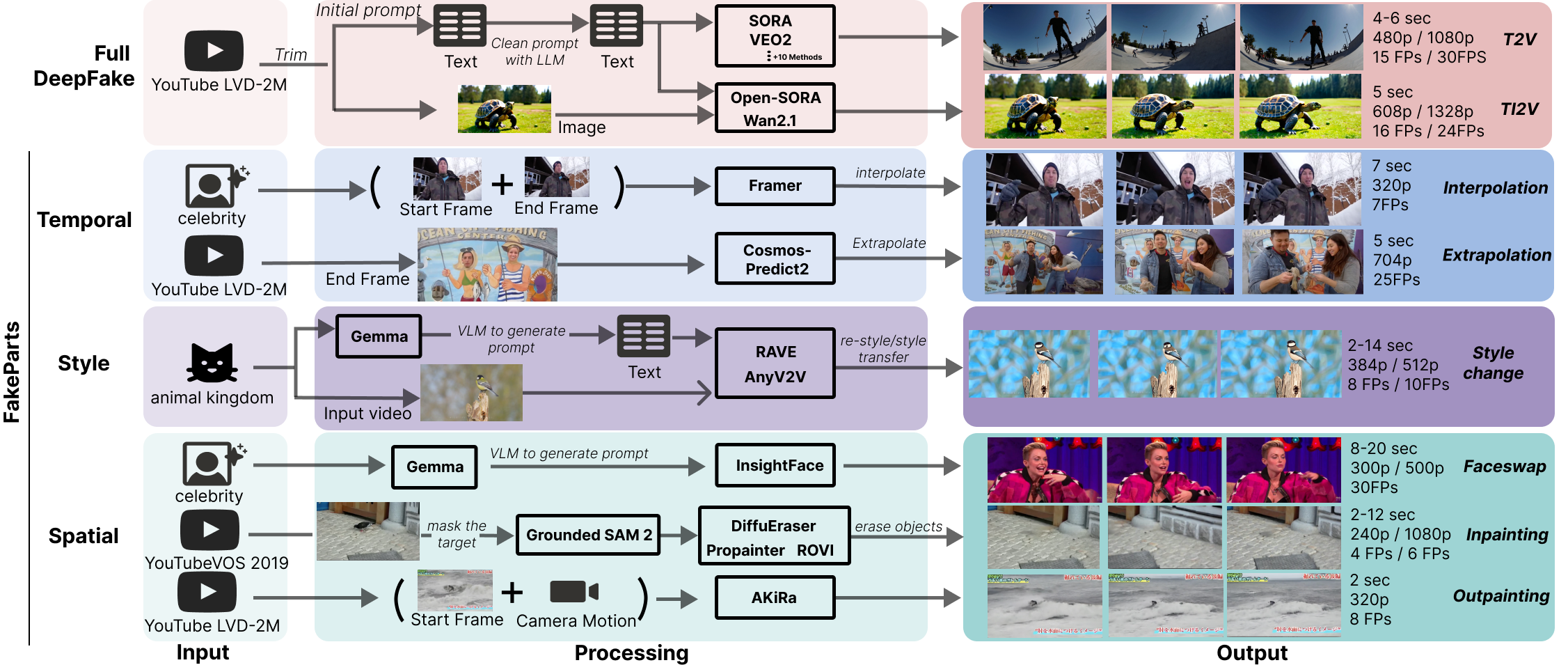}
  \caption{\textbf{Pipeline} of \datasetname{} includes both full and \deepfakename{} Deepfake videos. The partial manipulations, \deepfakename{}, are categorized into three types: temporal, spatial, and style. }  
  \label{fig:pipeline}
\end{figure*}

Figures~\ref{fig:by_wordcount}-\ref{fig:by_topic} present statistics with respect to prompt length (measured by word count) and video content topics.

\subsection{\datasetname{} generation process}
\label{sub:generation}

\begin{figure*}[ht]
    \centering
    \setlength{\tabcolsep}{4pt}
    \renewcommand{\arraystretch}{1.0}
    \begin{tabular}{cccc}
        \textbf{Faceswap} & \textbf{Inpainting} & \textbf{Interpolation} & \textbf{IT2V} \\
        \includegraphics[width=0.22\textwidth]{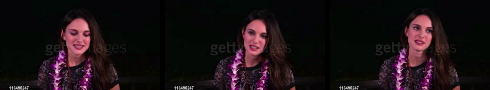} &
        \includegraphics[width=0.22\textwidth]{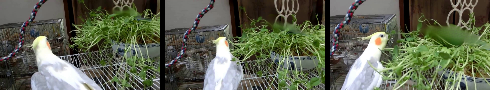} &
        \includegraphics[width=0.22\textwidth]{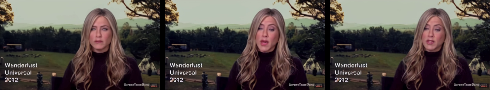} &
        \includegraphics[width=0.22\textwidth]{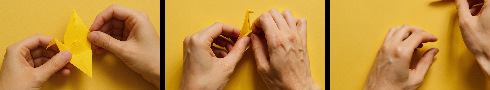} \\
        \includegraphics[width=0.22\textwidth]{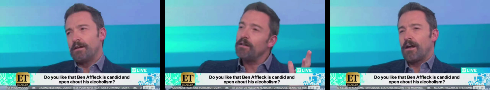} &
        \includegraphics[width=0.22\textwidth]{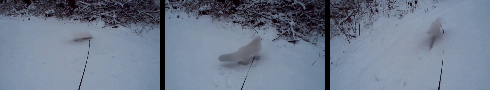} &
        \includegraphics[width=0.22\textwidth]{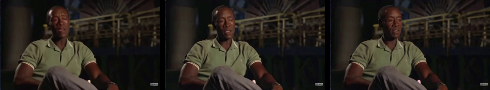} &
        \includegraphics[width=0.22\textwidth]{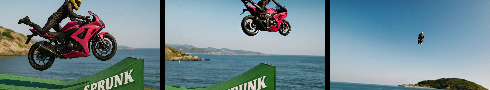} \\
        \includegraphics[width=0.22\textwidth]{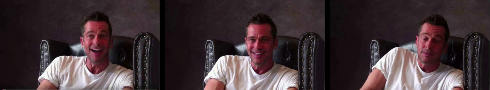} &
        \includegraphics[width=0.22\textwidth]{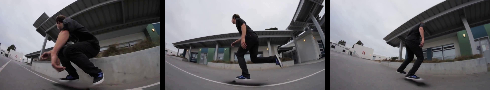} &
        \includegraphics[width=0.22\textwidth]{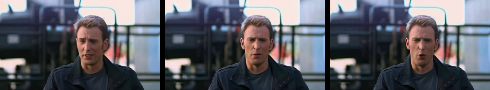} &
        \includegraphics[width=0.22\textwidth]{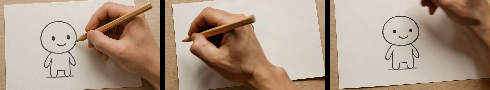} \\
        \textbf{Outpainting} & \textbf{Stylechange} & \textbf{T2V} & \textbf{Real} \\
        \includegraphics[width=0.22\textwidth]{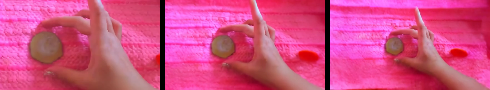} &
        \includegraphics[width=0.22\textwidth]{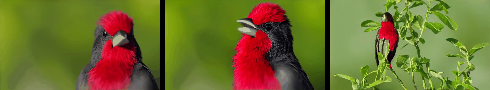} &
        \includegraphics[width=0.22\textwidth]{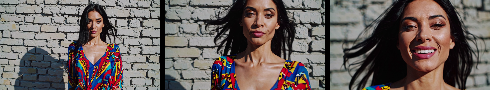} &
        \includegraphics[width=0.22\textwidth]{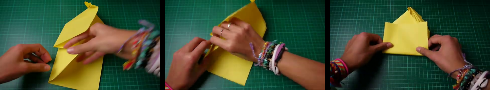} \\
        \includegraphics[width=0.22\textwidth]{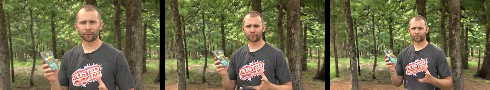} &
        \includegraphics[width=0.22\textwidth]{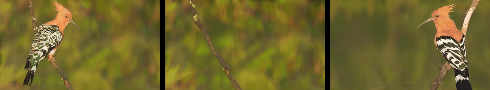} &
        \includegraphics[width=0.22\textwidth]{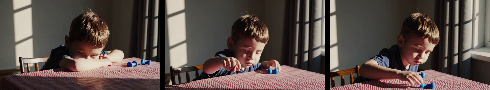} &
        \includegraphics[width=0.22\textwidth]{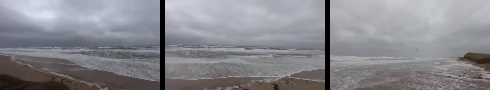} \\
        \includegraphics[width=0.22\textwidth]{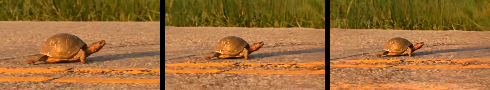} &
        \includegraphics[width=0.22\textwidth]{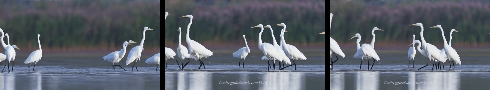} &
        \includegraphics[width=0.22\textwidth]{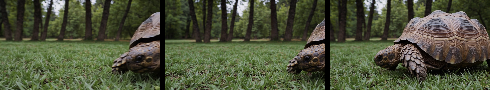} &
        \includegraphics[width=0.22\textwidth]{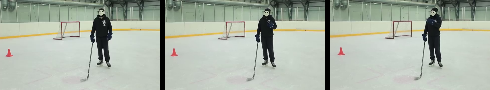} \\
    \end{tabular}
    \caption{\textbf{\deepfakename examples} grouped by manipulation group (per column) with three samples per group.}
    \label{fig:partial_deepfakes}
\end{figure*}

\datasetname{} includes both real and fake videos; below, we outline their collection and generation processes, also illustated in Figure~\ref{fig:pipeline}. Figure~\ref{fig:partial_deepfakes} shows some examples. 

\noindent \textbf{Real videos.} 
Many methods, particularly those involving partial manipulations, depend on real data. We use several publicly available datasets as real sources, including DAVIS 2016~\cite{Perazzi2016}, DAVIS 2017~\cite{pont20172017}, YouTube-VOS 2019~\cite{vos2019}, MOSE~\cite{MOSE} and LVD-2M~\cite{xiong2024lvd2m}. The specific usage of each dataset per method is detailed in the suppl.\ material. 

\noindent \textit{Vision-Language Models (VLMs).} Many videos are captioned or evaluated using VLMs. Unless stated otherwise, we use PaLI-Gemma 2 (3B) fine-tuned on VQAv2 ~\cite{steiner2024paligemma2familyversatile}.

\subsubsection{Full Deepfakes generation process} \label{paragraph: full deepfake}

The Full Deepfake category includes two types:  

First, Text-to-Video ({\color[rgb]{0.20,0.63,0.17}T2V}), which includes various open-source (CogVideoX~\cite{yang2025cogvideoxtexttovideodiffusionmodels}, 
Hunyuan Video~\cite{kong2025hunyuanvideosystematicframeworklarge}, 
LTX-Video~\cite{hacohen2024ltxvideorealtimevideolatent}, 
Mochi 1 by Genmo~\cite{genmo2024mochi}, 
Open-Sora~\cite{zheng2024opensorademocratizingefficientvideo}, 
Wan2.1~\cite{wan2025wanopenadvancedlargescale}, 
AllegroAI~\cite{zhou2024allegro}) and commercial (Sora~\cite{openai2024sora}, Veo2~\cite{google2025veo2}) generators published in 2024--2025. 
For all {\color[rgb]{0.20,0.63,0.17}T2V} methods, we generate diverse and semantically rich prompts using Google Gemini~\cite{geminiteam2025geminifamilyhighlycapable}.

Second, Text-and-Image-to-Video ({\color[rgb]{0.38,0.69,0.48}IT2V}), where we include the models that support dual-modals conditioning (text and image), i.e., Open-Sora~\cite{zheng2024opensorademocratizingefficientvideo}, Wan2.1~\cite{wan2025wanopenadvancedlargescale}. We use the Gemini VLM: 
We extract the first frame from real videos (\textit{e.g.}, Pexels dataset~\cite{McCallumCorranPexelvideos}) and query Gemini to produce a continuation prompt based on it. 
We will include the prompts and conditioning frames in our metadata. 

\begin{table*}[ht]
\centering

\small
\resizebox{0.92\textwidth}{!}{
\begin{tabular}{p{0.2\textwidth}@{} l l l p{0.6\textwidth}@{}}
\toprule
Category & Method & Venue & C &   Description \\
\midrule

\multirow{7}{*}{\parbox{0.17\textwidth}{{\color[rgb]{0.25,0.55,0.90}Out-of-the-box (\textit{OoB})} CNN- or ViT-based binary detectors trained on their respective datasets}}
& NPR~\cite{tan2024rethinking} & {\fontsize{7pt}{8pt}\selectfont\textit{CVPR'24}} 
 & I & {\fontsize{8pt}{10pt}\selectfont Focus on pixel-neighbour relations, local cues from upsampling, to capture general artefacts} \\
& HiFi~\cite{hifi_net_xiaoguo} & {\fontsize{7pt}{8pt}\selectfont\textit{CVPR'23}} & I & {\fontsize{8pt}{10pt}\selectfont Exploits hierarchical feature discrimination for GAN- and diffusion-generated images}   \\
& AIGVDet~\cite{bai2024ai} & {\fontsize{7pt}{8pt}\selectfont\textit{PRCV'24}} & V &  {\fontsize{8pt}{10pt}\selectfont Combines spatial and optical flow branches to detect frame-level inconsistencies, fused into robust video-level predictions} \\
& CNNDetection~\cite{wang2020cnn} & {\fontsize{7pt}{8pt}\selectfont\textit{CVPR'20}} 
  & I & {\fontsize{8pt}{10pt}\selectfont Trained on one generator (ProGAN), it generalizes by learning shared low-level artifacts} \\
& IML-ViT~\cite{ma_2024imlvit} & {\fontsize{7pt}{8pt}\selectfont\textit{NeurIPS'24}} 
  & I & {\fontsize{8pt}{10pt}\selectfont ViT-based Image Manipulation Localization method using high-res, multi-scale features and edge supervision to capture subtle traces} \\

\midrule
\multirow{6}{*}{\parbox{0.17\textwidth}{{\color[rgb]{0.90,0.35,0.25}Foundation-Model-Based (\textit{FoM})} detectors that leverage pretrained large-scale models (e.g., CLIP~\cite{radford2021learning})}}
& UnivFD~\cite{ojha2023towards} & {\fontsize{7pt}{8pt}\selectfont\textit{CVPR'23}} 
  & I & {\fontsize{8pt}{10pt}\selectfont CLIP embeddings with linear probes or nearest neighbours to detect fakes from GANs, diffusion, and autoregressive models} \\
& D3~\cite{yang2025d3} & {\fontsize{7pt}{8pt}\selectfont\textit{CVPR'25}} 
  & I & {\fontsize{8pt}{10pt}\selectfont Dual-branch CLIP-with-attention method that detects discrepancy artefacts}  \\
& C2P-CLIP~\cite{tan2025c2p} & {\fontsize{7pt}{8pt}\selectfont\textit{AAAI'25}} 
  & I & {\fontsize{8pt}{10pt}\selectfont CLIP-based detector that employs category–concept prompts} \\
& FatFormer~\cite{liu2024forgeryaware} & {\fontsize{7pt}{8pt}\selectfont\textit{CVPR'24}} & I & {\fontsize{8pt}{10pt}\selectfont CLIP backbone augmented with a forgery-aware adapter and prompt conditioning }
   \\
& De-Fake~\cite{sha2023defakedet} & {\fontsize{7pt}{8pt}\selectfont\textit{ArXiv'23}} 
  & I & {\fontsize{8pt}{10pt}\selectfont CLIP-aligned detector that decouples real/fake evidence; improves open-set generalisation} \\

\midrule
\multirow{6}{*}{\parbox{0.17\textwidth}{{\color[rgb]{0.80,0.45,0.75}VLM-Based (\textit{VLM})} use prompt- or fine-tuned VLMs that output predictions (no confidence scores)}}
& FakeVLM~\cite{wen2025spot} & {\fontsize{7pt}{8pt}\selectfont\textit{NeurIPS'25}} 
  &  I & {\fontsize{8pt}{10pt}\selectfont MLLM specialised for synthetic-image detection with natural-language artefact explanations}\\
& SIDA~\cite{DBLPHuangHLH00W0C25} & {\fontsize{7pt}{8pt}\selectfont\textit{CVPR'25}} 
  & I & {\fontsize{8pt}{10pt}\selectfont Extends a VLM with special tokens to capture image-level authenticity cues, enabling classification of real, synthetic, and tampered images}\\
& MM-Det~\cite{Song_2024_MM_Det} & {\fontsize{7pt}{8pt}\selectfont\textit{NeurIPS'24}} 
  &  V & {\fontsize{8pt}{10pt}\selectfont Integrates VLM (LLaVA) with spatio-temporal attention to detect diffusion-generated videos through joint visual–textual reasoning} \\
& AntifakePrompt~\cite{chang2024antifake} & {\fontsize{7pt}{8pt}\selectfont\textit{ArXiv'25}} 
  & I & {\fontsize{8pt}{10pt}\selectfont Prompt-tuned VLMs (InstructBLIP) cast as VQA; strong zero- and few-shot generalisation} \\

%%%
\midrule
\parbox{0.17\textwidth}{{\color[rgb]{0.30,0.75,0.85}Diffusion-Based (\textit{DiB})} use denoising signals from pretrained DM}
& DRCT~\cite{pmlr-v235-chen24ay} & {\fontsize{7pt}{8pt}\selectfont\textit{ICML'24}} 
  & I & {\fontsize{8pt}{10pt}\selectfont Leverages contrastive training on diffusion-inverted reconstructions to distinguish fake content from real videos} \\

% \end{itemize}
    
\bottomrule
\end{tabular}}
\caption{\textbf{Forgery methods evaluated.} Column `C' corresponds to image or video level category of method.}
\label{tab:methods_list}
\end{table*}

\subsubsection {\deepfakename{} generation process} 
\label{subsub:fakeparts}

\paragraph{I. Spatial \deepfakename} includes methods that alter regional information within the video content. 

\textit{{\color[rgb]{0.73,0.63,0.04}Faceswap}}. We replace the face in a target video with that from a source face image. For this, we use the a widely used tool for face detection and swapping, InsightFace library~\cite{ren2023pbidr, guo2021sample, gecer2021ostec, an_2022_pfc_cvpr, an_2021_pfc_iccvw, deng2020subcenter, Deng2020CVPR, guo2018stacked, deng2018menpo, deng2018arcface}. During generation, we use target videos from the Celeb-DF dataset~\cite{li2020celebdflargescalechallengingdataset} and source images from CelebA~\cite{liu2015faceattributes} to avoid the misuse of sensitive facial information. Both datasets focus exclusively on celebrities, whose facial identities are already publicly exposed. To ensure gender consistency after swapping, we prompt VLM with gender and filter the unmatched pairs.

\textit{{\color[rgb]{0.78,0.18,0.18}Inpainting} Generation.} 
We include two such sources.  
First, we generate inpainted videos using DiffuEraser~\cite{li2025diffueraserdiffusionmodelvideo} and ProPainter~\cite{zhou2023propainter}.  
The pipeline consists of: 
(1) extracting the first frame and using a VLM to identify a salient object to remove; 
(2) segmenting the object across all frames with Grounded-SAM-2~\cite{ravi2024sam2segmentimages}; and 
(3) applying the inpainting models to remove the object and fill the masked regions, producing realistic object-removal deepfakes.
Second, we include the inpainting dataset from ROVI~\cite{wu2024languagedrivenvideoinpaintingmultimodal}, where videos are inpainted using E2FGVI~\cite{li2022towards} with careful human curation and parameter tuning.

\textit{{\color[rgb]{0.42,0.42,0.42}Outpainting}.} %Unlike image~\cite{lugmayr2022repaint,anciukevivcius2023renderdiffusion}, 
Video outpainting accounts for camera motion and 3D consistency. Recent approaches address this through explicit camera control~\cite{xu2024camco,he2024cameractrl,wang2024motionctrl,wang2024akira}. We use AkiRA~\cite{wang2024akira}, built on SVD~\cite{blattmann2023stable}. For generation, we apply backwards camera tracking trajectories to guide coherent extrapolation at frame boundaries.

\vspace{2mm}
\noindent \textbf{II. Temporal \deepfakename} includes methods that complete or modify content along time axis, such as frame interpolation.

\textit{{\color[rgb]{1.00,0.68,0.26}Interpolation} Generation.} 
We employ Framer~\cite{wang2024framerinteractiveframeinterpolation} to synthesize smooth transitions between non-consecutive frames, simulating temporal edits. 
We randomly select 14 consecutive frames from each video, provide the first and last frames to Framer to generate 14 synthetic intermediate frames, and obtain a temporally edited video.

\textit{{\color[rgb]{0.73,0.63,0.04}Extrapolation} Generation.} 
We employ the Cosmos Predict Video2World framework~\cite{nvidia2025cosmospredict2github, nvidia2025cosmosworldfoundationmodel, nvidia2025worldsimulationvideofoundation} to generate temporally extrapolated videos from real footage. 
Given the final 50 frames of a captured sequence with their caption as prompt, the model predicts and synthesizes the subsequent motion and scene evolution, extending each clip to 120 frames. 
The resulting videos of 25FPS and 5 seconds, preserve the original scene context while introducing plausible forward dynamics, forming our extrapolation subset. 

\vspace{2mm}
\noindent \textbf{III. Style \deepfakename} targets manipulations that retain the semantic content of the video but modify its appearance through style transfer techniques.

% --- Add to your LaTeX preamble ---

\definecolor{lightgreen}{HTML}{C9EAD5}
\definecolor{lightred}{HTML}{F4C7C3}
% ---------------------------------

\begin{table*}[ht]
\centering
\caption{Performance comparison of video generators and detectors using \% ROC-AUC (or equivalent) scores. Green cells indicate scores above 85\%, while red cells denote scores below 53\%.}
\label{tab:performance_model_genre}
\small
\begin{tabular}{|c|l|r|c|c|c|c|c|c|c|c|}
\hline
Category & Gen vs. Det & Venue & {\color[rgb]{0.38,0.69,0.48}IT2V} & {\color[rgb]{0.20,0.63,0.17}T2V} & {\color[rgb]{1.00,0.68,0.26}Interp.} & {\color[rgb]{0.73,0.63,0.04}Extrap.} & {\color[rgb]{0.78,0.18,0.18}Inpaint.} & {\color[rgb]{0.42,0.42,0.42}Outpaint.} & {\color[rgb]{0.73,0.63,0.04}Face} & {{\color[rgb]{0.55,0.47,0.72}Style}} \\
\hline
- & \textbf{Random} & -- 
  & 50.0 & 50.0 & 50.0 & 50.0 & 50.0 & 50.0 & 50.0 & 50.0 \\
\hline
\multirow{5}{*}{{\color[rgb]{0.25,0.55,0.90}OoB}}
& NPR~\cite{tan2024rethinking} & {\fontsize{7pt}{8pt}\selectfont\textit{CVPR24'}} 
  & 68.3 & 72.8 & 57.0 & 61.1 & \cellcolor{lightred}37.2 & 59.2 & 66.7 & 59.5 \\
& HiFi~\cite{hifi_net_xiaoguo} & {\fontsize{7pt}{8pt}\selectfont\textit{CVPR23'}} 
  & 60.3 & 60.1 & 53.7 & \cellcolor{lightred}45.8 & \cellcolor{lightred}45.5 & \cellcolor{lightred}49.9 & 65.2 & 65.2 \\
& AIGV~\cite{bai2024ai} & {\fontsize{7pt}{8pt}\selectfont\textit{PRCV24'}} 
  & \cellcolor{lightred}48.0 & \cellcolor{lightred}46.3 & \cellcolor{lightred}48.1 & 51.4 & \cellcolor{lightred}37.3 & 57.6 & 82.2 & \cellcolor{lightred}33.4 \\
& CNNDet~\cite{wang2020cnn} & {\fontsize{7pt}{8pt}\selectfont\textit{CVPR20'}} 
  & 55.5 & \cellcolor{lightgreen}94.8 & \cellcolor{lightred}54.0 & 62.3 & 67.8 & 57.4 & \cellcolor{lightred}45.9 & 66.1 \\
& IML-ViT~\cite{ma_2024imlvit} & {\fontsize{7pt}{8pt}\selectfont\textit{NeurIPS24'}} 
  & \cellcolor{lightred}32.8 & 61.2 & 57.2 & \cellcolor{lightred}45.3 & 56.0 & \cellcolor{lightred}44.1 & 58.2 & \cellcolor{lightred}34.0 \\
\hline
\multirow{5}{*}{{\color[rgb]{0.90,0.35,0.25}FoM}}
& UnivFD~\cite{ojha2023towards} & {\fontsize{7pt}{8pt}\selectfont\textit{CVPR23'}} 
  & 65.0 & \cellcolor{lightgreen}89.1 & 65.9 & 58.1 & 83.1 & 57.9 & 58.4 & 83.5 \\
& D3~\cite{yang2025d3} & {\fontsize{7pt}{8pt}\selectfont\textit{CVPR25'}} 
  & 81.8 & \cellcolor{lightgreen}94.8 & 75.4 & 70.6 & 72.9 & 71.3 & \cellcolor{lightred}37.5 & 83.5 \\
& C2P~\cite{tan2025c2p} & {\fontsize{7pt}{8pt}\selectfont\textit{AAAI25'}} 
  & 83.1 & \cellcolor{lightgreen}96.5 & \cellcolor{lightgreen}88.9 & 84.3 & 68.3 & 82.2 & 60.6 & \cellcolor{lightgreen}88.1 \\
& Fatfrm.~\cite{liu2024forgeryaware} & {\fontsize{7pt}{8pt}\selectfont\textit{CVPR24'}} 
  & \cellcolor{lightred}43.2 & 63.1 & \cellcolor{lightred}50.2 & \cellcolor{lightred}47.8 & 53.7 & 53.2 & \cellcolor{lightred}47.7 & 56.8 \\
& De-Fake~\cite{sha2023defakedet} & {\fontsize{7pt}{8pt}\selectfont\textit{ArXiv23'}} 
  & \cellcolor{lightgreen}94.4 & 70.9 & \cellcolor{lightred}48.0 & \cellcolor{lightred}49.2 & \cellcolor{lightred}44.4 & \cellcolor{lightgreen}96.3 & \cellcolor{lightred}48.8 & \cellcolor{lightred}49.0 \\
\hline
\multirow{4}{*}{{\color[rgb]{0.80,0.45,0.75}VLM}}
& FakeVLM~\cite{wen2025spot} & {\fontsize{7pt}{8pt}\selectfont\textit{NeurIPS25'}} 
  & \cellcolor{lightred}46.3 & \cellcolor{lightred}47.2 & \cellcolor{lightred}50.6 & \cellcolor{lightred}50.4 & \cellcolor{lightred}46.4 & \cellcolor{lightred}51.3 & \cellcolor{lightred}46.2 & 54.3 \\
& SIDA~\cite{DBLPHuangHLH00W0C25} & {\fontsize{7pt}{8pt}\selectfont\textit{CVPR25'}} 
  & 83.9 & \cellcolor{lightred}52.6 & \cellcolor{lightred}47.2 & \cellcolor{lightred}46.3 & \cellcolor{lightred}34.5 & \cellcolor{lightred}47.3 & 62.3 & \cellcolor{lightred}25.4 \\
& MM-Det~\cite{Song_2024_MM_Det} & {\fontsize{7pt}{8pt}\selectfont\textit{NeurIPS24'}} 
  & 54.3 & \cellcolor{lightred}52.8 & 61.9 & 53.5 & \cellcolor{lightred}37.1 & 59.9 & 63.1 & \cellcolor{lightred}52.7 \\
& AFPrompt~\cite{chang2024antifake} & {\fontsize{7pt}{8pt}\selectfont\textit{ArXiv25'}} 
  & \cellcolor{lightred}50.0 & \cellcolor{lightred}50.0 & \cellcolor{lightred}50.0 & \cellcolor{lightred}50.0 & \cellcolor{lightred}50.0 & \cellcolor{lightred}50.0 & \cellcolor{lightred}50.0 & \cellcolor{lightred}50.0 \\
\hline
{\color[rgb]{0.30,0.75,0.85}DiB}
& DRCT~\cite{pmlr-v235-chen24ay} & {\fontsize{7pt}{8pt}\selectfont\textit{ICML24'}} 
  & 72.0 & \cellcolor{lightgreen}86.0 & \cellcolor{lightgreen}97.2 & \cellcolor{lightred}44.6 & 70.9 & \cellcolor{lightgreen}95.2 & 71.9 & \cellcolor{lightgreen}96.4 \\
\hline
- & \textbf{Human} & -- & 83.1 & \cellcolor{lightgreen}86.3 & 60.2 & 60.0 & 83.3 & 83.3 & 80.0 & 82.0 \\
\hline
\end{tabular}
\end{table*}

\textit{{{\color[rgb]{0.55,0.47,0.72}Style Change}} Generation.} We use RAVE~\cite{kara2023raverandomizednoiseshuffling} to alter color and appearance on videos from Animal Kingdom~\cite{Ng_2022_CVPR}. A random frame is captioned with PaLI-Gemma2~\cite{steiner2024paligemma2familyversatile}, and RAVE is prompted with: \textit{`Change the colour of the animal in the video.'}
We also include object-specific colour edits using AnyV2V~\cite{ku2024anyv2v}, where we caption frames using VideoChat-R1~\cite{li2025videochatr1enhancingspatiotemporalperception}, and then extract the object to be edited and pair it with target colour. The resulting prompt (\textit{`Change the colour of the car to blue while keeping the rest of the scene unchanged.'}) ensures only local edits. 
\section{Experimental Protocol}
\label{sec:experiments}

We benchmark state-of-the-art image and video detectors on \datasetname{}. Specifically, all detectors are grouped into four groups according to their underlying detection mechanisms, as explained in Table~\ref{tab:methods_list}: {{\color[rgb]{0.25,0.55,0.90}Out-of-the-box (\textit{OoB})}, {{\color[rgb]{0.90,0.35,0.25}Foundation-Model-Based (\textit{FoM})}, {{\color[rgb]{0.80,0.45,0.75}VLM-Based (\textit{VLM})}, and {{\color[rgb]{0.30,0.75,0.85}Diffusion-Based (\textit{DiB})}. 

Each video is decoded into full-resolution without any resizing, aspect-ratio normalisation, or geometric alignment. This ensures that the frames retain their original aspect ratio and any artefacts introduced during generation and preserves the native spatial characteristics of the data, allowing subsequent analysis on unaltered visual content rather than resized or warped patches.

\begin{figure*}[htp!]
  \centering
  \includegraphics[width=1.0\linewidth]{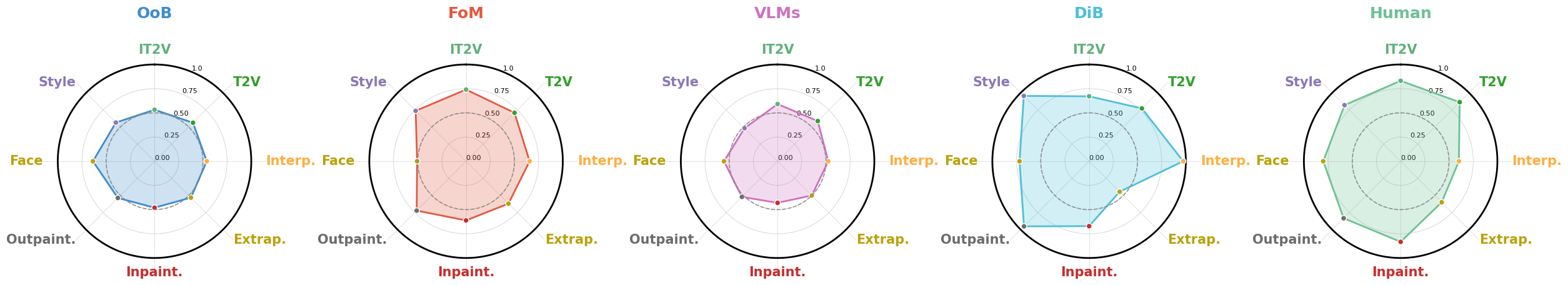}
  \caption{\textbf{Results} in Radar map of \datasetname{} performance across detector categories, where for each category we report the average performance achieved among all generators:
{\color[rgb]{0.25,0.55,0.90}{OoB}}, 
{\color[rgb]{0.90,0.35,0.25}{FoM}}, 
{\color[rgb]{0.80,0.45,0.75}{VLM}}, and 
{\color[rgb]{0.30,0.75,0.85}{DiB}} and
on various generators, including \textbf{full fake} generators such as: {\color[rgb]{0.20,0.63,0.17}{T2V}}, 
{\color[rgb]{0.38,0.69,0.48}{TI2V}}, and \deepfakename{}: 
{\color[rgb]{0.55,0.47,0.72}{Change-of-Style}}, 
{\color[rgb]{0.78,0.18,0.18}{Inpainting}}, 
{\color[rgb]{0.42,0.42,0.42}{Outpainting}}, 
{\color[rgb]{1.00,0.68,0.26}{Interpolation}}, 
{\color[rgb]{0.73,0.63,0.04}{Extrapolation}}, and 
{\color[rgb]{0.73,0.63,0.04}{Faceswap}}.}

\label{fig:radar method}
\end{figure*}

\subsection{Detection Performance}

Table~\ref{tab:performance_model_genre} reports the AUC-ROC scores per model, when available, given that some VLM-based detectors output direct binary predictions. 
Each row corresponds to a detector grouped into:
{\color[rgb]{0.25,0.55,0.90}Out-of-the-Box (OoB\textbf{)}},
{\color[rgb]{0.90,0.35,0.25}Foundation-Model-Based (FoM)},
{\color[rgb]{0.80,0.45,0.75}Vision–Language–Model-Based (VLM)}, and
{\color[rgb]{0.30,0.75,0.85}Diffusion-Based (DiB)}.
Columns report detector performance across generator families, including
\textbf{Full-Fake Generators}
({\color[rgb]{0.20,0.63,0.17}T2V} and
{\color[rgb]{0.38,0.69,0.48}TI2V})
and \textbf{Fake-Part Generators}
({\color[rgb]{0.55,0.47,0.72}Change-of-Style},
{\color[rgb]{0.78,0.18,0.18}Inpainting},
{\color[rgb]{0.42,0.42,0.42}Outpainting},
{\color[rgb]{1.00,0.68,0.26}Interpolation},
{\color[rgb]{0.73,0.63,0.04}Extrapolation}, and
{\color[rgb]{0.73,0.63,0.04}Faceswap}).
Figure~\ref{fig:radar method} shows category-wise performances, aggregating the detectors within each category into radar plots that reveal consistent trends across generator types. 
We draw the following findings:

\begin{tcolorbox}[colback=white,colframe=black!70!white,boxrule=0.5pt,left=2pt,right=2pt,top=1pt,bottom=1pt]
\textbf{1. Full-fake content is comparatively easy to detect.}  
\end{tcolorbox} \vspace{-2mm}
\noindent Across nearly all detectors, full-fakes such as  
{\color[rgb]{0.38,0.69,0.48}\textit{IT2V}} and  
{\color[rgb]{0.20,0.63,0.17}\textit{T2V}}  
are detected with relatively high and consistent score (more green cells in Table~\ref{tab:performance_model_genre}).  
For example, {\color[rgb]{0.20,0.63,0.17}\textit{T2V}} videos, which contain no real frames, are detected by most FoM detectors, the diffusion-based detector (DiB), and even human raters at above 85\% AUC. Interestingly, introducing only \textit{a single real frame}, as in  
{\color[rgb]{0.38,0.69,0.48}\textit{IT2V}}, already reduces performance compared to pure  
{\color[rgb]{0.20,0.63,0.17}\textit{T2V}}: C2P~\cite{tan2025c2p} reduces 96.5\% to 83.1\% and DRCT~\cite{pmlr-v235-chen24ay} reduces from 86\% to 72\%.
Moreover, detector performance drops substantially for \deepfakename{}, as seen in the columns for  
{\color[rgb]{0.78,0.18,0.18}\textit{Inpaint.}},  
{\color[rgb]{1.00,0.68,0.26}\textit{Interp.}}, and  
{\color[rgb]{0.73,0.63,0.04}\textit{Extrap.}}.  
Many detectors exhibit weak performance (red cells), with some AUCs even falling below 50\%, i.e., worse than random chance. 

\begin{tcolorbox}[colback=white,colframe=black!70!white,boxrule=0.5pt,left=2pt,right=2pt,top=1pt,bottom=1pt]
\textbf{2. OoB and VLM-based detectors perform poorly.} 
\end{tcolorbox}   \vspace{-2mm}
\noindent OoB CNN/ViT-style detectors and VLM-based approaches often operate near or below random-chance accuracy, particularly on \deepfakename{} generators such as  
{\color[rgb]{0.78,0.18,0.18}\textit{Inpaint.}}.  
Although they occasionally achieve reasonable performance on certain full-fake settings; e.g., CNNDet~\cite{wang2020cnn} reaches 94.8\% AUC on  
{\color[rgb]{0.20,0.63,0.17}\textit{T2V}}, and SIDA~\cite{DBLPHuangHLH00W0C25} 83.9\% on {\color[rgb]{0.38,0.69,0.48}\textit{IT2V}}. Their behavior is unstable across manipulation types. Both drop to below $\sim$65\% AUC for nearly all other generator.
% , highlighting the lack of robustness of OoB and VLM-based detectors when confronted with diverse or localized fake content.

\begin{tcolorbox}[colback=white,colframe=black!70!white,boxrule=0.5pt,left=2pt,right=2pt,top=1pt,bottom=1pt]
\textbf{3. FoM detectors work well on full-fake content but are brittle on \deepfakename{}.}  
\end{tcolorbox}   \vspace{-2mm}
\noindent FoM achieve strong performance on full-fake videos:  
{\color[rgb]{0.38,0.69,0.48}\textit{IT2V}} (e.g., De-Fake~\cite{sha2023defakedet} reaches 94.4\%) and  
{\color[rgb]{0.20,0.63,0.17}\textit{T2V}} (e.g., C2P~\cite{tan2025c2p} at 96.5\%, D3 at 94.8\%, UnivFD at 89.1\%).
However, their performance drops sharply on \deepfakename{} manipulations such as  
{\color[rgb]{0.78,0.18,0.18}\textit{Inpaint.}} (C2P at 68.3\%),  
{\color[rgb]{0.42,0.42,0.42}\textit{Outpaint.}} (UnivFD at 57.9\%),  
{\color[rgb]{1.00,0.68,0.26}\textit{Interp.}} (De-Fake at only 48.0\%), and  
{\color[rgb]{0.73,0.63,0.04}\textit{Face}} (none of the FoM methods exceed 60\%). 
Notably, most FoM detectors rely on CLIP and almost none handle inpainting, face swap, or extrapolation reliably, revealing a core weakness of current CLIP-based approaches.

\begin{tcolorbox}[colback=white,colframe=black!70!white,boxrule=0.5pt,left=2pt,right=2pt,top=1pt,bottom=1pt]
\textbf{4. Diffusion-based detection offers the best balance between accuracy and coverage.}  
\end{tcolorbox}   \vspace{-2mm}
\noindent Compared with others the diffusion-based model exhibits the most balanced performance profile in both Table~\ref{tab:performance_model_genre} and Figure~\ref{fig:radar method}. It maintains strong accuracy across full-fake  
{\color[rgb]{0.38,0.69,0.48}\textit{IT2V}} and  
{\color[rgb]{0.20,0.63,0.17}\textit{T2V}}  
as well as across diverse \deepfakename{} manipulations, with the only notable weakness being the extrapolation case (44.6\%).  
Although it does not always achieve the single best score on every axis, its combination of high accuracy and broad robustness makes diffusion-based detection the most reliable category overall in our benchmark.

\begin{tcolorbox}[colback=white,colframe=black!70!white,boxrule=0.5pt,left=2pt,right=2pt,top=1pt,bottom=1pt]
\textbf{5. \deepfakename{} impact human performance.}
\end{tcolorbox} \vspace{-2mm}
\noindent Comparing detector performance with human judgments reveals several informative parallels.  
Humans achieve their highest AUC (86.3\%) on full-fake videos, particularly  
{\color[rgb]{0.20,0.63,0.17}\textit{T2V}},  
mirroring the performance trends observed in many detectors.  
For \deepfakename{} manipulations, human evaluators remain relatively sensitive to  
{\color[rgb]{0.78,0.18,0.18}\textit{Inpaint.}},  
{\color[rgb]{0.42,0.42,0.42}\textit{Outpaint.}},  
and {\color[rgb]{0.73,0.63,0.04}Face}, typically achieving AUC values above 80\%.  
However, they perform significantly worse on  
{\color[rgb]{1.00,0.68,0.26}\textit{Interp.}}  
and  
{\color[rgb]{0.73,0.63,0.04}\textit{Extrap.}},  
both around 60\%.  
We believe this pattern reflects inherent differences in human sensitivity to spatial versus temporal inconsistencies: spatial artifacts (e.g., inpainting) are easier to perceive, whereas temporal irregularities (e.g., interpolation) are harder to differentiate. Interestingly, while humans are often weaker than the top detectors on full-fake content, they remain the \textit{most balanced and reliable} judges for \deepfakename{} content, as shown on the right side of Figure~\ref{fig:radar method}. This further highlights the challenge that partial edits pose for current automated detection systems.

\begin{figure}[ht!]
\begin{minipage}[b]{0.46\linewidth}
    \includegraphics[width=0.95\textwidth]{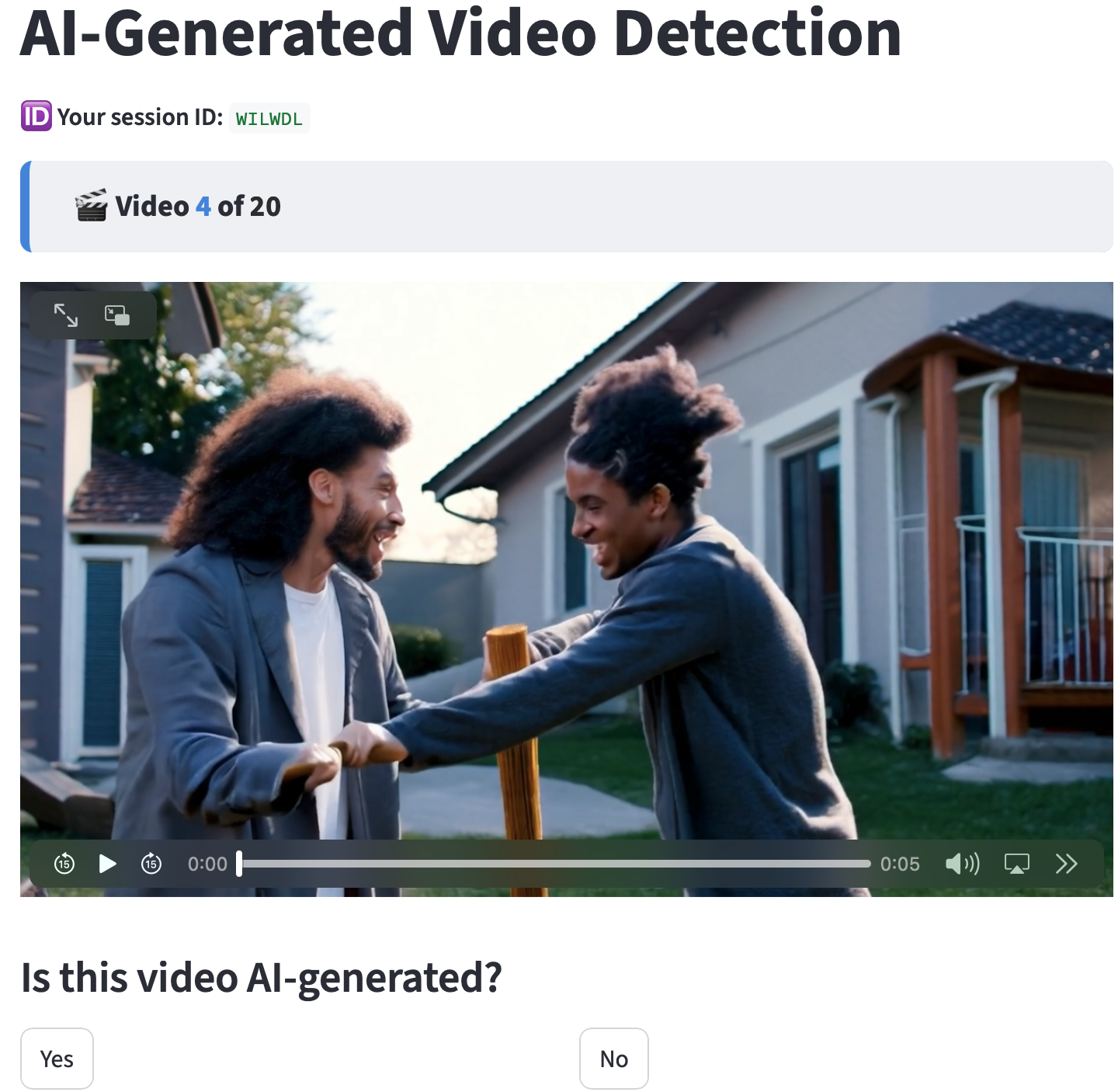}
\end{minipage}
\hfill
\begin{minipage}[b]{0.46\linewidth}
   \includegraphics[width=0.95\textwidth]{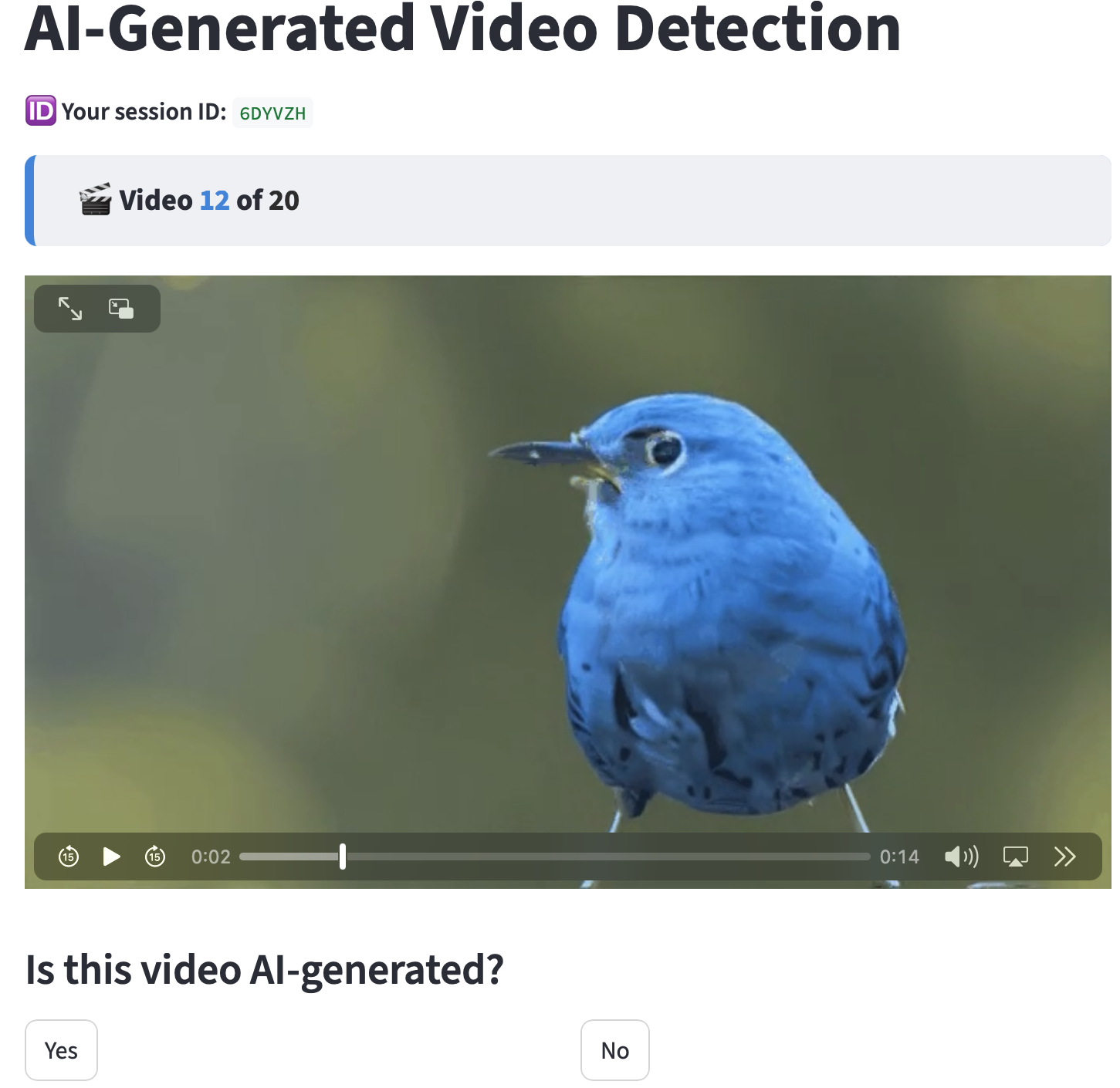}
\end{minipage}
\caption{\textbf{GUI for user study:} Participants classify videos as real or fake, we show  full-fake (left) and \deepfakename{} (right) examples. } 
\label{fig:Annotationtool}
\end{figure}

% %%%%%%%%%%%%%

\noindent \textbf{$\to$ Why do these patterns emerge?}  
We attribute the weaknesses of OoB and VLM detectors to domain shift. OoB detectors rely heavily on fingerprints and low-level artifacts~\cite{bammey2023synthbuster}, which are inconsistent across modern generators, while VLMs depend strongly on their fine-tuning distribution and thus fail to generalize to synthetic, stylized, or partially edited content. FoM detectors, many of which are CLIP-based, leverage semantic-level representations, enabling good generalization across full-fake videos and some \deepfakename{}. However, semantic or SSL-style features tend to overlook small, localized edits; real pixels can dilute forged regions. A clear example is  
{\color[rgb]{0.73,0.63,0.04}\textit{Face}},  
where edit areas are small and no FoM exceeds 60\% AUC. By contrast, diffusion-based detectors achieve the best balance as their reconstruction-based mechanism evaluates whether inputs lie on the manifold of real data, rather than depending on semantic cues or brittle low-level fingerprints~\cite{pmlr-v235-chen24ay,wang2024your,li2023your}. This allows them to capture subtle artifacts in both full-fake and \deepfakename, resulting in uniformly strong performances.

\subsection{Human Perception Study}

To assess human detection capabilities, we conducted a user study using a Streamlit-based survey. %\footnote{\url{https://genaidetection.streamlit.app/}}. 
A total of 290 participants each labelled 20 randomly selected video clips (real or \deepfakename{}), yielding approximately 5800 annotations overall, half corresponding to real content and half to videos containing \emph{FakeParts}. Humans achieved an average of 77.3\% accuracy, outperforming most automated methods, especially on partial fakes \textit{Interpolation}, \textit{Extrapolation}, \textit{Inpainting}, \textit{Outpainting}, \textit{Faceswap} and \textit{Style Change}. 
\section{Conclusion}

We introduced \textbf{\datasetname}, the first benchmark for detecting \textit{\deepfakename}, deepfakes with localized manipulations in otherwise authentic videos. These subtle edits exploit real context, severely reducing the performance of both human annotators and state-of-the-art detectors. By providing fine-grained spatial and temporal annotations, \datasetname{} enables rigorous evaluation under partial manipulations. We hope it drives progress toward detectors that capture fine-grained inconsistencies and uphold media integrity against evolving generative manipulations.

\vspace{2mm} \noindent \textbf{Limitations.}
 While we cover diverse manipulation types, finer-grained exploration, such as varying inpainting areas and their effect on detection, remains open. Moreover, although our goal is to strengthen detection, we recognize potential risks of misuse; our focus remains strictly on defense and mitigation for trustworthy AI.
\section*{Acknowledgements} 
This work is supported by Hi!PARIS and ANR/France 2030 program (ANR-23-IACL-0005) and ANR APATE (ANR-22-CE39-0016), CIEDS vague 4, Hi!Paris grant, postdoctoral fellowship and Professor Monge chair. 

This project was granted access to the IDRIS High-Performance Computing (HPC) resources under the allocation 20XX-AD011015141R2 made by GENCI (Grand Équipement National de Calcul Intensif). We sincerely thank Eleftherios Tsonis for his meticulous proofreading.
{
    \small
    \bibliographystyle{ieeenat_fullname}
    \bibliography{main}
}

\newpage 
\appendix
% \startcontents[sections] 
% \printcontents[sections]{l}{1}{\setcounter{tocdepth}{3}} 
\appendix

\section{Detailed Experiment Results}

\paragraph{More detailed experiment results}
We present in Figure~\ref{fig:sup_generator_method} and Table~\ref{tab:performance_model_gen_det_sup} the detailed statistics of our benchmark.  
We observe the following trends:  
(1) All {\color[rgb]{0.25,0.55,0.90}{OoB}} detectors exhibit consistently low performance, covering only a small area in the radar plots.  
(2) {\color[rgb]{0.90,0.35,0.25}{FoM}} methods demonstrate large performance variance: C2P~\cite{tan2025c2p} and D3~\cite{yang2025d3} perform very well, but predominantly on the {\color[rgb]{0.38,0.69,0.48}{Full-fake}} region (right half of the radar map). Both models, however, fail notably on {\color[rgb]{0.78,0.18,0.18}{Inpainting}}, especially the ROVI variant.  
(3) Within {\color[rgb]{0.80,0.45,0.75}{VLM}}-based detectors, only SIDA~\cite{DBLPHuangHLH00W0C25} shows partial effectiveness, again mainly on {\color[rgb]{0.38,0.69,0.48}{Full-fake}} generators.  
(4) For {\color[rgb]{0.30,0.75,0.85}{DiB}}, DRCT~\cite{pmlr-v235-chen24ay} achieves the most balanced performance across both {\color[rgb]{0.38,0.69,0.48}{Full-fake}} and fake-part categories, performing strongly on {\color[rgb]{0.78,0.18,0.18}{Inpainting}}, {\color[rgb]{0.55,0.47,0.72}{Change-of-Style}}, and several {\color[rgb]{0.73,0.63,0.04}{Faceswap}} cases. However, its notably weak result on {\color[rgb]{0.73,0.63,0.04}{Extrapolation}} (Cosmos), together with the fact that it rarely exceeds the very-high-performance range (above 85\%), suggests that diffusion-based detectors still struggle to reach high-accuracy regimes. This indicates that even the best-performing DiB method leaves the overall detection problem posed by the \datasetname{} benchmark far from solved.
% }

\paragraph{Analysis of the Most and Least Detectable Samples.}
Figure~\ref{fig:sup_generator_method} further highlights the challenge of detecting \deepfakename{} manipulations by visualizing the distribution of the most- and least-detectable samples. The most-detected group (i.e., videos flagged by at least 12 detectors) is dominated by Full-Fake content---especially {\color[rgb]{0.20,0.63,0.17}\textit{T2V}}---confirming that fully synthetic videos remain comparatively easy for detectors across all categories. In contrast, the least-detected samples show a strikingly different pattern: almost all are detected exclusively by DRCT~\cite{pmlr-v235-chen24ay} and missed entirely by every other detector. This reinforces our earlier finding that DRCT is the most balanced detector in the benchmark, capable of capturing subtle inconsistencies overlooked by others. Interestingly, these difficult cases span both Full-Fake and \deepfakename{} examples, indicating that partial and localized manipulations continue to pose substantial challenges for most existing detectors. Figure~\ref{fig:extreme_examples} presents qualitative examples of both extremes. We see that many of the most-detected samples appear visually convincing, even to human observers, suggesting that strong detector performance does not necessarily correlate with human perceptual difficulty. This divergence between machine and human sensitivity underscores the need for deeper understanding of what cues different detectors rely on, especially in the presence of \deepfakename{} artifacts.

\begin{figure*}[htp!]
  \centering
  \includegraphics[width=1.0\linewidth]{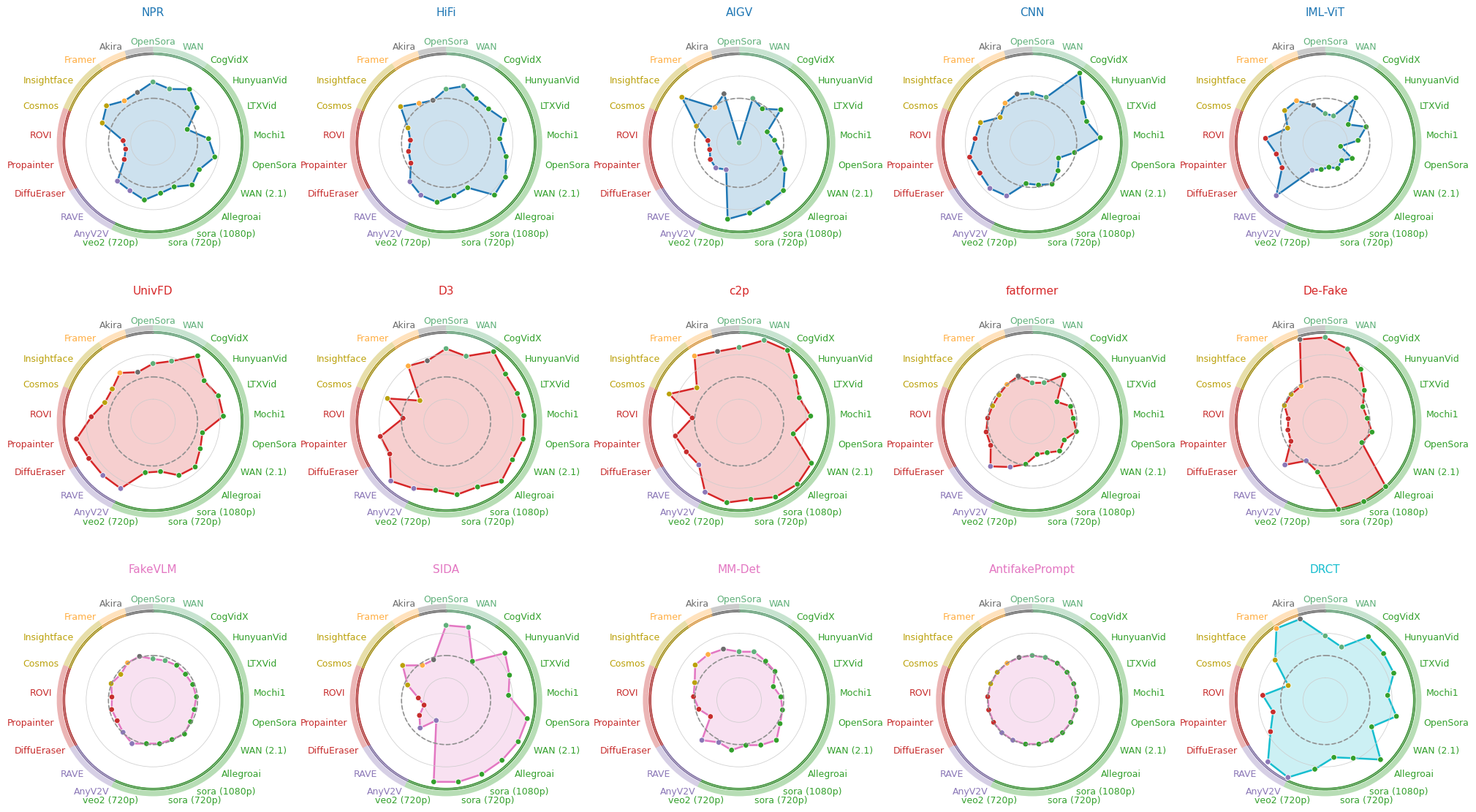}
  \caption{\textbf{Results} in Radar map of \datasetname{} performance across detector categories, where for each category we report the average performance achieved among all generators:
{\color[rgb]{0.25,0.55,0.90}{OoB}}, 
{\color[rgb]{0.90,0.35,0.25}{FoM}}, 
{\color[rgb]{0.80,0.45,0.75}{VLM}}, and 
{\color[rgb]{0.30,0.75,0.85}{DiB}} and
on various generators, including \textbf{full fake} generators such as: {\color[rgb]{0.20,0.63,0.17}{T2V}}, 
{\color[rgb]{0.38,0.69,0.48}{TI2V}}, and \deepfakename{}: 
{\color[rgb]{0.55,0.47,0.72}{Change-of-Style}}, 
{\color[rgb]{0.78,0.18,0.18}{Inpainting}}, 
{\color[rgb]{0.42,0.42,0.42}{Outpainting}}, 
{\color[rgb]{1.00,0.68,0.26}{Interpolation}}, 
{\color[rgb]{0.73,0.63,0.04}{Extrapolation}}, and 
{\color[rgb]{0.73,0.63,0.04}{Faceswap}}.}
\label{fig:sup_generator_method}
\end{figure*}

\section{Generated Data Summary in \datasetname}

The table~\ref{tab:detailed_video_counts} summarises the detailed statistics of both the generated and real data included in \datasetname{}. 
It provides an overview of all 21 methods covered in our benchmark, including 
\textit{Style Change}~\cite{ku2024anyv2v, 
kara2023raverandomizednoiseshuffling}, 
\textit{Extrapolation}~\cite{nvidia2025cosmospredict2github, nvidia2025cosmosworldfoundationmodel}, 
\textit{FaceSwap}~\cite{deng2018arcface}, 
\textit{Inpainting}~\cite{li2025diffueraserdiffusionmodelvideo, zhou2023propainter, wu2024languagedrivenvideoinpaintingmultimodal}, 
\textit{Interpolation}~\cite{wang2024framerinteractiveframeinterpolation}, 
\textit{Outpainting}~\cite{wang2024akira}, 
\textit{T2V}~\cite{yang2025cogvideoxtexttovideodiffusionmodels, kong2025hunyuanvideosystematicframeworklarge, hacohen2024ltxvideorealtimevideolatent, genmo2024mochi, zheng2024opensorademocratizingefficientvideo, wan2025wanopenadvancedlargescale, zhou2024allegro, openai2024sora, google2025veo2}, 
and \textit{TI2V}~\cite{zheng2024opensorademocratizingefficientvideo, wan2025wanopenadvancedlargescale}.
For each method, we list the number of real and fake videos, the corresponding frame counts, and the native resolution, frame rate, and clip duration (or a range when applicable).

\begin{table*}[ht]
\centering
\caption{
Performance comparison of video generators and detectors in terms of ROC-AUC (in \%). 
Cells in green indicate scores $\ge 85\%$, while red cells indicate scores $\le 53\%$. 
Detector columns are color-coded by family: 
{\color[rgb]{0.25,0.55,0.90}{OoB}}, 
{\color[rgb]{0.90,0.35,0.25}{FoM}}, 
{\color[rgb]{0.80,0.45,0.75}{VLM}}, and 
{\color[rgb]{0.30,0.75,0.85}{DiB}}. 
Generator rows are grouped by genre, including \textbf{full-fake} generators 
{\color[rgb]{0.20,0.63,0.17}{T2V}} and 
{\color[rgb]{0.38,0.69,0.48}{TI2V}}, 
as well as \deepfakename{} manipulations such as 
{\color[rgb]{0.55,0.47,0.72}{Change-of-Style}}, 
{\color[rgb]{0.78,0.18,0.18}{Inpainting}}, 
{\color[rgb]{0.42,0.42,0.42}{Outpainting}}, 
{\color[rgb]{1.00,0.68,0.26}{Interpolation}}, 
{\color[rgb]{0.73,0.63,0.04}{Extrapolation}}, and 
{\color[rgb]{0.73,0.63,0.04}{Faceswap}}.
}
\label{tab:performance_model_gen_det_sup}
% \tiny
\resizebox{0.95\linewidth}{!}{%
\begin{tabular}{|l|c|c|c|c|c|c|c|c|c|c|c|c|c|c|c|}
\hline
Generator 
& {\color[rgb]{0.25,0.55,0.90}NPR}
& {\color[rgb]{0.25,0.55,0.90}HiFi}
& {\color[rgb]{0.25,0.55,0.90}AIGV}
& {\color[rgb]{0.25,0.55,0.90}CNNDet}
& {\color[rgb]{0.25,0.55,0.90}IML-ViT}
& {\color[rgb]{0.90,0.35,0.25}UnivFD}
& {\color[rgb]{0.90,0.35,0.25}D3}
& {\color[rgb]{0.90,0.35,0.25}C2P}
& {\color[rgb]{0.90,0.35,0.25}Fatfrm.}
& {\color[rgb]{0.90,0.35,0.25}De-Fake}
& {\color[rgb]{0.80,0.45,0.75}FakeVLM}
& {\color[rgb]{0.80,0.45,0.75}SIDA}
& {\color[rgb]{0.80,0.45,0.75}MM-Det}
& {\color[rgb]{0.80,0.45,0.75}AFPrompt}
& {\color[rgb]{0.30,0.75,0.85}DRCT} \\
\hline
\textbf{Random} 
& 50.0 & 50.0 & 50.0 & 50.0 
& 50.0 & 50.0 & 50.0 & 50.0 
& 50.0 & 50.0 & 50.0 & 50.0 
& 50.0 & 50.0 & 50.0 \\
\hline
{\color[rgb]{0.38,0.69,0.48}\textbf{OpenSora-768p}} % TI2V
& 62.9 & 57.9 & --   & 56.0 & \cellcolor{lightred}38.9 & 63.6 & 80.6 & 83.1 
& \cellcolor{lightred}45.3 & \cellcolor{lightgreen}94.4 & \cellcolor{lightred}49.9 & \cellcolor{lightgreen}86.3 
& \cellcolor{lightred}52.4 & \cellcolor{lightred}50.0 & 69.5 \\
{\color[rgb]{0.38,0.69,0.48}\textbf{WAN}} % TI2V
& 59.5 & 63.2 & 57.1 & 54.0 & \cellcolor{lightred}38.6 & 69.1 & 76.1 & \cellcolor{lightgreen}95.3 
& \cellcolor{lightred}47.4 & 84.9 & \cellcolor{lightred}49.9 & \cellcolor{lightgreen}86.9 
& 53.8 & \cellcolor{lightred}50.0 & 65.2 \\
{\color[rgb]{0.20,0.63,0.17}\textbf{CogVidX}} % T2V
& 66.4 & 63.7 & \cellcolor{lightred}52.2 & \cellcolor{lightgreen}94.5 & 55.2 & \cellcolor{lightgreen}86.3 
& \cellcolor{lightgreen}91.0 & \cellcolor{lightgreen}97.3 & 57.9 & 68.4 & \cellcolor{lightred}49.4 
& 53.8 & \cellcolor{lightred}51.4 & \cellcolor{lightred}50.0 & 83.0 \\
{\color[rgb]{0.20,0.63,0.17}\textbf{HunyuanVid}} % T2V
& 59.4 & 56.2 & 68.3 & 75.9 & \cellcolor{lightred}40.8 & 74.2 & 83.1 & 84.1 
& \cellcolor{lightred}42.1 & 57.5 & \cellcolor{lightred}49.7 & \cellcolor{lightgreen}85.3 
& \cellcolor{lightred}50.9 & \cellcolor{lightred}50.0 & 84.0 \\
{\color[rgb]{0.20,0.63,0.17}\textbf{LTXVid}} % T2V
& \cellcolor{lightred}42.9 & 70.9 & \cellcolor{lightred}44.6 & 64.4 & \cellcolor{lightred}48.4 & 76.1 
& \cellcolor{lightgreen}85.8 & 77.1 & \cellcolor{lightred}47.4 & \cellcolor{lightred}46.7 
& \cellcolor{lightred}49.3 & 76.0 & \cellcolor{lightred}45.2 & \cellcolor{lightred}50.0 & 83.0 \\
{\color[rgb]{0.20,0.63,0.17}\textbf{Mochi1}} % T2V
& 59.0 & 57.8 & \cellcolor{lightred}45.8 & 73.6 & \cellcolor{lightred}41.0 & 79.0 
& \cellcolor{lightgreen}86.6 & 83.2 & \cellcolor{lightred}47.5 & \cellcolor{lightred}47.9 
& \cellcolor{lightred}49.5 & 74.0 & \cellcolor{lightred}48.0 & \cellcolor{lightred}50.0 & 66.8 \\
{\color[rgb]{0.20,0.63,0.17}\textbf{OpenSora}} % T2V
& 64.5 & 64.4 & 56.5 & \cellcolor{lightred}49.0 & \cellcolor{lightred}34.4 & 56.6 
& \cellcolor{lightgreen}88.6 & 63.4 & \cellcolor{lightred}49.5 & 53.6 
& \cellcolor{lightred}49.4 & \cellcolor{lightgreen}94.3 & \cellcolor{lightred}49.8 
& \cellcolor{lightred}50.0 & 81.1 \\
{\color[rgb]{0.20,0.63,0.17}\textbf{WAN (2.1)}} % T2V
& 56.2 & 74.2 & 63.0 & \cellcolor{lightred}39.3 & \cellcolor{lightred}39.7 & 58.9 
& \cellcolor{lightgreen}86.5 & \cellcolor{lightgreen}93.7 & \cellcolor{lightred}44.3 
& \cellcolor{lightred}48.1 & \cellcolor{lightred}49.4 & \cellcolor{lightgreen}94.0 
& \cellcolor{lightred}51.7 & \cellcolor{lightred}50.0 & 63.5 \\
{\color[rgb]{0.20,0.63,0.17}\textbf{Allegroai}} % T2V
& 59.6 & 76.9 & 71.8 & \cellcolor{lightred}43.9 & \cellcolor{lightred}36.7 & 67.6 
& \cellcolor{lightgreen}91.5 & \cellcolor{lightgreen}96.5 & \cellcolor{lightred}47.2 
& \cellcolor{lightgreen}99.8 & \cellcolor{lightred}51.2 & \cellcolor{lightgreen}92.9 
& 56.6 & \cellcolor{lightred}50.0 & \cellcolor{lightgreen}91.9 \\
{\color[rgb]{0.20,0.63,0.17}\textbf{sora (1080p)}} % T2V
& 53.2 & \cellcolor{lightred}52.9 & 74.7 & \cellcolor{lightred}51.0 & \cellcolor{lightred}38.4 & 65.2 
& 81.7 & \cellcolor{lightgreen}94.5 & \cellcolor{lightred}42.6 
& \cellcolor{lightgreen}99.8 & \cellcolor{lightred}49.6 & \cellcolor{lightgreen}93.2 
& 53.5 & \cellcolor{lightred}50.0 & 73.8 \\
{\color[rgb]{0.20,0.63,0.17}\textbf{sora (720p)}} % T2V
& 54.7 & 56.6 & 79.9 & \cellcolor{lightred}48.4 & \cellcolor{lightred}36.9 & 55.5 
& 82.9 & \cellcolor{lightgreen}89.3 & \cellcolor{lightred}41.8 
& \cellcolor{lightgreen}99.7 & \cellcolor{lightred}49.8 & \cellcolor{lightgreen}93.6 
& \cellcolor{lightred}50.7 & \cellcolor{lightred}50.0 & 69.1 \\
{\color[rgb]{0.20,0.63,0.17}\textbf{veo2 (720p)}} % T2V
& 60.0 & 63.8 & \cellcolor{lightgreen}86.2 & \cellcolor{lightred}45.1 & \cellcolor{lightred}38.0 
& 56.8 & 79.1 & \cellcolor{lightgreen}92.2 & \cellcolor{lightred}49.3 
& 65.1 & \cellcolor{lightred}49.8 & \cellcolor{lightgreen}93.4 
& 53.9 & \cellcolor{lightred}50.0 & 75.1 \\
{\color[rgb]{0.55,0.47,0.72}\textbf{AnyV2V}} % Style
& 56.6 & 61.4 & \cellcolor{lightred}43.6 & 64.5 & \cellcolor{lightred}40.0 & 83.2 
& \cellcolor{lightgreen}85.2 & \cellcolor{lightgreen}90.7 & 56.3 
& \cellcolor{lightred}49.0 & \cellcolor{lightred}52.9 & \cellcolor{lightred}38.2 
& \cellcolor{lightred}51.3 & \cellcolor{lightred}50.0 & \cellcolor{lightgreen}96.3 \\
{\color[rgb]{0.55,0.47,0.72}\textbf{RAVE}} % Style
& 55.7 & 57.5 & \cellcolor{lightred}41.9 & 65.5 & 82.1 & 81.0 
& \cellcolor{lightgreen}89.4 & 73.7 & 64.5 & 66.1 & \cellcolor{lightred}49.7 
& \cellcolor{lightred}49.1 & 56.6 & \cellcolor{lightred}50.0 & \cellcolor{lightgreen}95.1 \\
{\color[rgb]{0.78,0.18,0.18}\textbf{DiffuEraser}} % Inpaint.
& \cellcolor{lightred}41.0 & \cellcolor{lightred}46.7 & \cellcolor{lightred}46.8 & 65.5 & 53.7 & 80.6 
& 72.7 & 72.6 & \cellcolor{lightred}51.6 & \cellcolor{lightred}45.1 
& \cellcolor{lightred}49.8 & \cellcolor{lightred}39.9 
& \cellcolor{lightred}42.7 & \cellcolor{lightred}50.0 & 68.4 \\
{\color[rgb]{0.78,0.18,0.18}\textbf{Propainter}} % Inpaint.
& \cellcolor{lightred}38.1 & \cellcolor{lightred}45.2 & \cellcolor{lightred}42.7 & 71.8 & 53.8 
& \cellcolor{lightgreen}88.5 & 76.3 & 79.4 & \cellcolor{lightred}51.0 
& \cellcolor{lightred}44.2 & \cellcolor{lightred}49.3 & \cellcolor{lightred}36.2 
& \cellcolor{lightred}47.9 & \cellcolor{lightred}50.0 & 56.6 \\
{\color[rgb]{0.78,0.18,0.18}\textbf{ROVI}} % Inpaint.
& \cellcolor{lightred}39.5 & \cellcolor{lightred}42.7 & \cellcolor{lightred}44.4 & 62.8 & 61.3 
& 68.6 & \cellcolor{lightred}50.4 & 58.5 & \cellcolor{lightred}48.7 
& \cellcolor{lightred}43.8 & \cellcolor{lightred}49.9 & \cellcolor{lightred}38.2 
& \cellcolor{lightred}50.8 & \cellcolor{lightred}50.0 & 68.2 \\
{\color[rgb]{0.73,0.63,0.04}\textbf{Cosmos}} % Extrap.
& 75.6 & 65.7 & 55.3 & 76.9 & 67.5 & 75.5 & 84.4 & \cellcolor{lightgreen}91.7 
& 67.9 & 68.9 & 69.3 & 66.7 & \cellcolor{lightred}51.9 & 69.1 & 81.9 \\
{\color[rgb]{0.73,0.63,0.04}\textbf{Insightface}} % Faceswap
& 68.3 & 65.4 & 84.6 & 53.0 & 62.8 & 60.5 & \cellcolor{lightred}51.2 & 64.6 
& 58.0 & 61.4 & 58.1 & 66.0 & 57.5 & 58.2 & 65.9 \\
{\color[rgb]{1.00,0.68,0.26}\textbf{Framer}} % Interp.
& \cellcolor{lightred}34.3 & \cellcolor{lightred}32.3 & \cellcolor{lightred}50.1 & \cellcolor{lightred}32.9 
& \cellcolor{lightred}34.8 & \cellcolor{lightred}45.6 & 60.1 & 81.1 
& \cellcolor{lightred}30.5 & \cellcolor{lightred}29.2 & \cellcolor{lightred}30.7 
& \cellcolor{lightred}29.2 & 56.9 & \cellcolor{lightred}30.4 & \cellcolor{lightgreen}95.3 \\
{\color[rgb]{0.42,0.42,0.42}\textbf{Akira}} % Outpaint.
& 56.1 & \cellcolor{lightred}49.2 & 65.3 & 55.2 & \cellcolor{lightred}45.9 & 56.4 
& 71.9 & 84.3 & \cellcolor{lightred}52.3 & \cellcolor{lightgreen}96.7 
& \cellcolor{lightred}50.7 & \cellcolor{lightred}48.4 & 55.6 
& \cellcolor{lightred}50.0 & \cellcolor{lightgreen}95.1 \\
\hline
\end{tabular}
}
\end{table*}

\section{Extra information on the Human Perception Study}\label{sec:humanstudy_sup}

To rigorously assess the perceptual quality of the generated deepfake videos in our new benchmark, we developed a dedicated annotation platform to collect large-scale human judgments.
This platform enables participants to watch each video and indicate whether they believe it is real or fake.
After submitting their judgment, they are prompted to provide a short explanation justifying their decision.
% An illustration of the interface can be seen in Figure \ref{fig:Annotationtool_1}.
In total, \textbf{290} individuals participated in the annotation process, providing both binary classifications and qualitative feedback.Annotators interacted with a simple, intuitive UI that ensures both accessibility and consistency across evaluations.
Metadata such as response times and annotator confidence levels were not collected to keep the privacy of annotators.
After analysing the qualitative responses, we identified four recurring types of perceptual cues commonly cited by annotators when labelling videos as fake:

\begin{figure*}[htp!]
  \centering
  \includegraphics[width=1.0\linewidth]{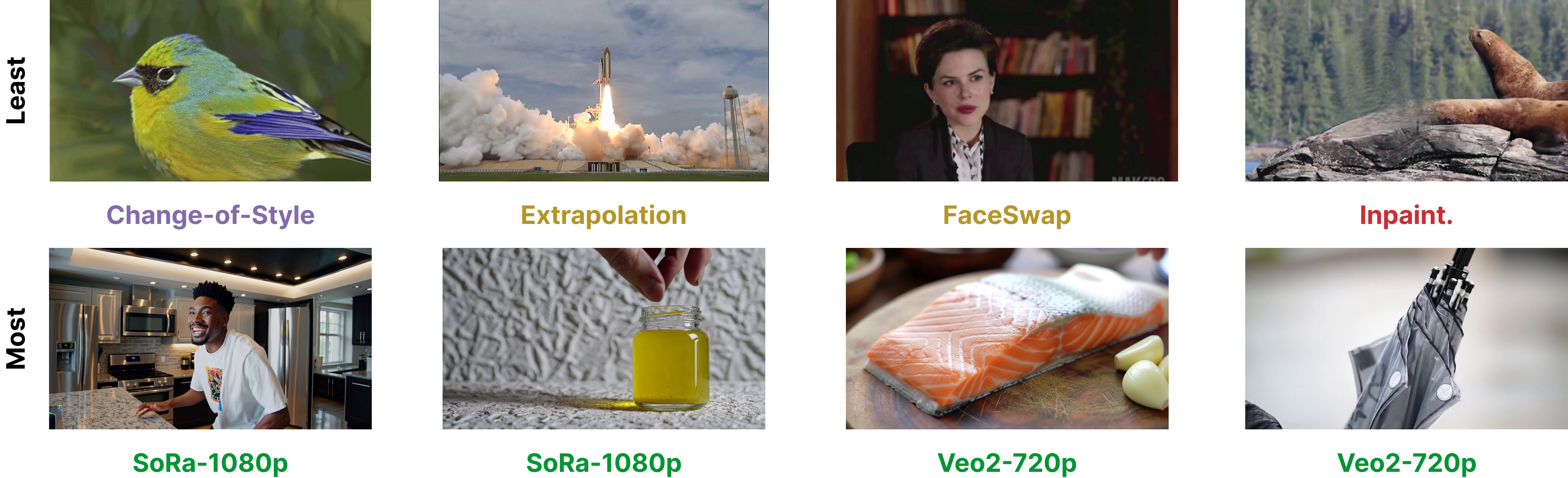}
  \caption{\textbf{Demonstrations} of the least and most detectable generated samples.  
The \textbf{first row} shows four \textit{least detected} examples, identified \textbf{only} by DRCT~\cite{pmlr-v235-chen24ay} and missed by all other detectors.  
The \textbf{second row} presents the \textit{most detected} examples, each correctly flagged by \textbf{more than 12 detectors}. A clear pattern emerges: highly detectable cases are predominantly \textbf{full-fake videos}, whereas the most challenging cases are almost exclusively \textbf{\deepfakename{}} samples.}
\label{fig:extreme_examples}
\end{figure*}

\begin{figure*}[htp!]
  \centering
  \includegraphics[width=1.0\linewidth]{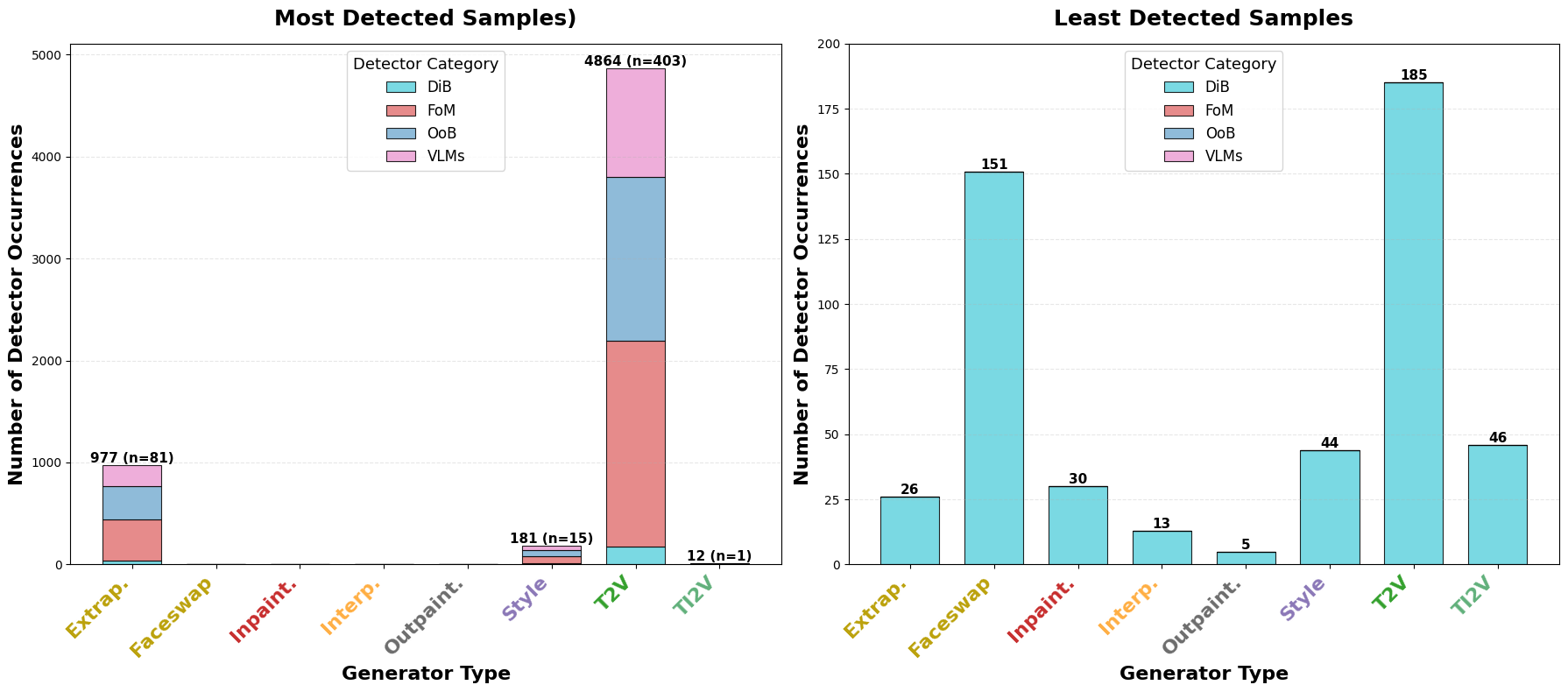}
  \caption{\textbf{Histogram} of the least and most detectable generated samples. 
We sample 500 of the most-detected and 500 of the least-detected videos and collect their occurrences (note that we have 15 detectors). 
For the most-detected group, the vast majority are Full-Fakes, especially {\color[rgb]{0.20,0.63,0.17}{T2V}} generations. 
In contrast, the least-detected samples are almost exclusively detected by a single detector, DRCT~\cite{pmlr-v235-chen24ay}, and missed by all others, consistent with our observation that DRCT is the most balanced detector. 
Moreover, these least-detected examples span both Full-Fakes and {\deepfakename{}}, further highlighting our finding that partial manipulations are significantly harder for most detectors.}
\label{fig:extreme_examples_histo}
\end{figure*}

% Use in Appendix
\begin{table*}[htp!]
    \centering
    \small
    \renewcommand{\arraystretch}{1.5}
    \setlength{\tabcolsep}{4pt}
    \caption{Detailed statistics for all methods.\label{tab:detailed_video_counts}}
    \resizebox{\linewidth}{!}{
    \begin{tabular}{
        l l l l
        r r r
        r r r
        l c c
    }
        \toprule
        \textbf{Year} & \textbf{Venue} & \textbf{Task} & \textbf{Method} &
        \textbf{RealV} & \textbf{FakeV} & \textbf{TotV} &
        \textbf{RealF} & \textbf{FakeF} & \textbf{TotF} &
        \textbf{Resolutions} & \textbf{FPS} & \textbf{Dur} \\
        \midrule
        % Style changes
        2024 & TMLR 2024 & Style Change & AnyV2V~\cite{ku2024anyv2v}
        & 0 & 1,545 & 1,545 & 0 & 24,720 & 24,720
        & 512$\times$512 & 8 & 2 \\
        2023 & CVPR 2024 & Style Change & RAVE~\cite{kara2023raverandomizednoiseshuffling}
        & 0 & 3,721 & 3,721 & 0 & 487,476 & 487,476
        & 680$\times$384 & 10 & 14 \\
        % Extrapolations
        2025 & arXiv & Extrapolation & Cosmos-Predict2~\cite{nvidia2025cosmospredict2github}
        & 120,000 & 3,000 & 3,000 & 120,000 & 268,618 & 388,618
        & 1280$\times$704 & 25 & 5 \\
        % Face swap
        2019 & CVPR 2019 & Faceswap & InsightFace~\cite{deng2018arcface}
        & 0 & 5,036 & 5,036 & 0 & 1,990,470 & 1,990,470
        & --- & 30 & 11 \\
        % inpaintings
        2025 & arXiv & Inpainting & DiffuEraser~\cite{li2025diffueraserdiffusionmodelvideo}
        & 0 & 2,243 & 2,243 & 0 & 75,871 & 75,871
        & 1280$\times$720 & 6 & 6 \\
        2023 & ICCV 2023 & Inpainting & ProPainter~\cite{zhou2023propainter}
        & 0 & 2,244 & 2,244 & 0 & 75,907 & 75,907
        & 1280$\times$720 & 6 & 6 \\
        2024 & CVPR 2024 & Inpainting & ROVI~\cite{wu2024languagedrivenvideoinpaintingmultimodal}
        & 7,252 & 9,090 & 9,090 & 16,342 & 199,032 & 247,014
        & 426$\times$320--1280$\times$720 & 5 & 4 \\
        % 2024 & CVPR 2024 & Inpainting & ROVI-REAL\_JPEG\_MP4~\cite{wu2024languagedrivenvideoinpaintingmultimodal}
        % & 7,252 & 0 & 7,252 & 199,032 & 0 & 199,032
        % & 426$\times$320--1280$\times$720 & 5 & 4 \\
        % interpolations
        2025 & ICLR 2025 & Interpolation & Framer~\cite{wang2024framerinteractiveframeinterpolation}
        & 320,000 & 10,000 & 10,000 & 320,000 & 140,000 & 460,000
        & 512$\times$320 & 7 & 7 \\
        % outpaintings
        % 2025 & CVPR 2025 & Outpainting & akira~\cite{wang2024akira}
        % & 0 & 2,000 & 2,000 & 0 & 28,000 & 28,000
        % & 576$\times$320 & 8 & 2 \\
        2025 & CVPR 2025 & Outpainting & Akira~\cite{wang2024akira}
        & 0 & 7,275 & 7,275 & 0 & 101,850 & 101,850
        & 576$\times$320 & 8 & 2 \\
        % t2v
        2025 & ICLR & T2V & CogVideoX~\cite{yang2025cogvideoxtexttovideodiffusionmodels}
        & 0 & 2,720 & 2,720 & 0 & 377,629 & 377,629
        & 720$\times$480 & 25 & 5 \\
        2024 & arXiv & T2V & HunyuanVideo~\cite{kong2025hunyuanvideosystematicframeworklarge}
        & 0 & 1,550 & 1,550 & 0 & 109,646 & 109,646
        & 1280$\times$720 & 15 & 4 \\
        2024 & arXiv & T2V & LTXVideo~\cite{hacohen2024ltxvideorealtimevideolatent}
        & 0 & 2,605 & 2,605 & 0 & 365,629 & 365,629
        & 1280$\times$720 & 30 & 5 \\
        2024 & Report & T2V & Mochi1~\cite{genmo2024mochi}
        & 0 & 1,059 & 1,059 & 0 & 137,367 & 137,367
        & 848$\times$480 & 25 & 5 \\
        2024 & arXiv & T2V & Open-Sora~\cite{zheng2024opensorademocratizingefficientvideo}
        & 0 & 2,000 & 2,000 & 0 & 258,000 & 258,000
        & 1024$\times$576 & 24 & 5 \\
        2025 & arXiv & T2V & Wan (2.1)~\cite{wan2025wanopenadvancedlargescale}
        & 0 & 1,423 & 1,423 & 0 & 115,263 & 115,263
        & 1280$\times$720 & 16 & 5 \\
        2024 & arXiv & T2V & allegroai~\cite{zhou2024allegro}
        & 0 & 1,000 & 1,000 & 0 & 88,000 & 88,000
        & 1280$\times$720 & 15 & 6 \\
        2025 & OpenAI Report & T2V & sora-1080p~\cite{openai2024sora}
        & 0 & 997 & 997 & 0 & 148,871 & 148,871
        & 1920$\times$1080 & 30 & 5 \\
        2025 & OpenAI Report & T2V & sora-720p~\cite{openai2024sora}
        & 0 & 1,766 & 1,766 & 0 & 264,900 & 264,900
        & 1280$\times$720 & 30 & 5 \\
        2024 & Google Report & T2V & veo2-720p~\cite{google2025veo2}
        & 0 & 999 & 999 & 0 & 119,880 & 119,880
        & 1280$\times$720 & 24 & 5 \\
        % ti2v
        2024 & arXiv & TI2V & Open-Sora-768px~\cite{zheng2024opensorademocratizingefficientvideo}
        & 0 & 1,980 & 1,980 & 0 & 255,420 & 255,420
        & 1024$\times$576 & 24 & 5 \\
        2025 & arXiv & TI2V & WAN~\cite{wan2025wanopenadvancedlargescale}
        & 0 & 2,000 & 2,000 & 0 & 162,000 & 162,000
        & 720$\times$1280 & 16 & 5 \\
        % Reals
        -- & -- & Real & TenKReal
        & 10,000 & 0 & 10,000 & 1,428,515 & 0 & 1,428,515
        & 640$\times$360 & 30 & 6 \\
        \midrule
        \textbf{ALL} & \textbf{--} & \textbf{--} & \textbf{--}
        & \textbf{17,252} & \textbf{64,252} & \textbf{81,504}
        & \textbf{2,067,507} & \textbf{5,814,442} & \textbf{7,881,949}
        & \textbf{--} & \textbf{--} & \textbf{--} \\
        \bottomrule
    \end{tabular}
    }
\end{table*}
\paragraph{Human Rationales.}

\begin{figure*}[htp!]
    \centering
    % first sub-figure
    \begin{subfigure}[t]{0.54\textwidth}
        \centering
        \includegraphics[width=\textwidth]{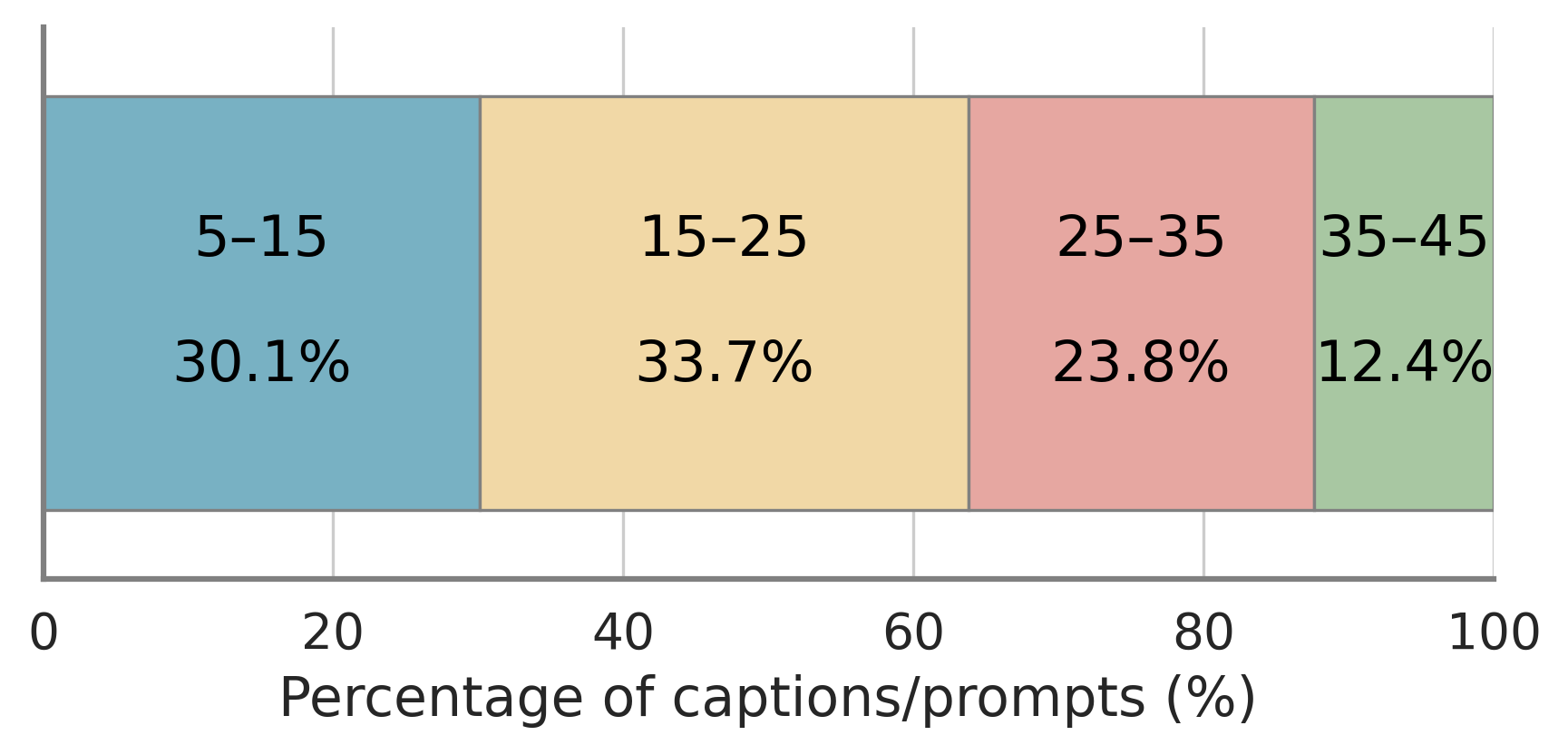}
        \caption{Word-count distribution of \datasetname{}}
        \label{fig:by_wordcount_3}
    \end{subfigure}
    \hfill
    % second sub-figure
    \begin{subfigure}[t]{0.44\textwidth}
        \centering
        \includegraphics[width=\textwidth]{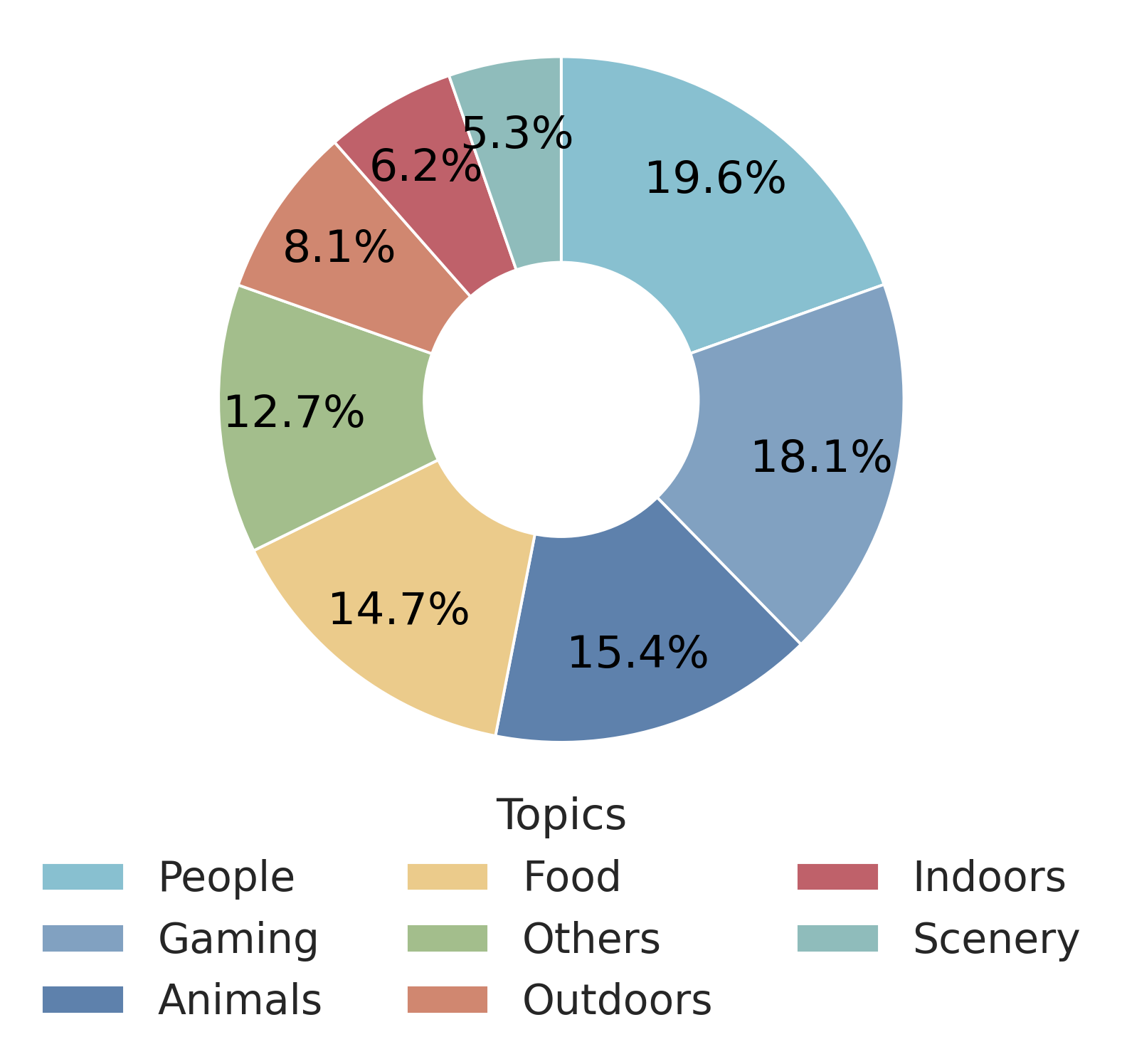}
        \caption{Topic distribution of \datasetname{}}
        \label{fig:by_topic_length}
    \end{subfigure}

    \caption{The distribution in terms of the caption length and theme category of the generated video content in our \datasetname{}.}% \TODO{This needs to be updated.}}
    \label{fig:combined_distributions_sup}
\end{figure*}

We analyzed participant comments and found four recurring patterns:

\begin{enumerate}
    \item \textbf{Temporal Inconsistencies:} unnatural transitions, jitter, or overly smooth motion.
    \item \textbf{Facial and Body Artifacts:} misaligned expressions, inconsistent eye/lip movement, weird blending.
    \item \textbf{Texture and Lighting Cues:} blurry details, mismatched lighting/shadows.
    \item \textbf{Semantic Anomalies:} unrealistic object behavior, physics violations.
\end{enumerate}

\begin{figure*}[htp!]
    \centering
    \includegraphics[width=0.6\textwidth]{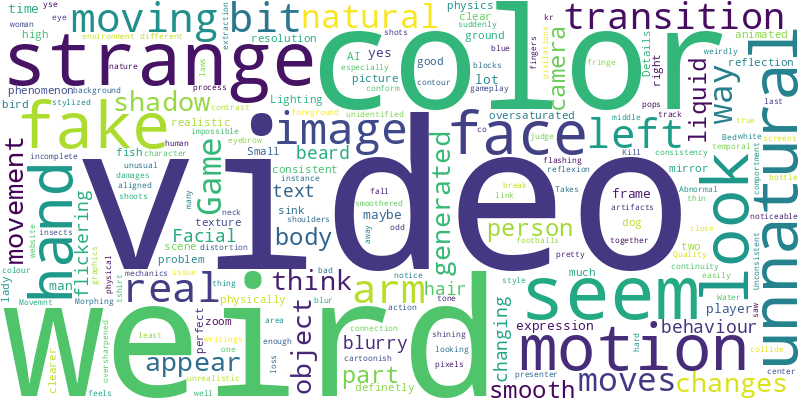}
    \caption{Word cloud showing the most frequent terms from human explanations of synthetic content detection.}
    \label{fig:wordcloud_sup}
\end{figure*}

\paragraph{Implications.}
Human perception is strongly influenced by low-level visual cues and spatio-temporal coherence.
To evade detection, generative models must improve:
\begin{itemize}
    \item \textit{Motion realism} (physically plausible dynamics),
    \item \textit{Temporal consistency} (stable shadows and lighting),
    \item \textit{High-frequency detail} (sharp but coherent textures).
\end{itemize}

\section{Implementation details}
\label{sec:benchmark_details_sup}

\begin{figure*}
    \centering
    \includegraphics[width=1.0\textwidth]{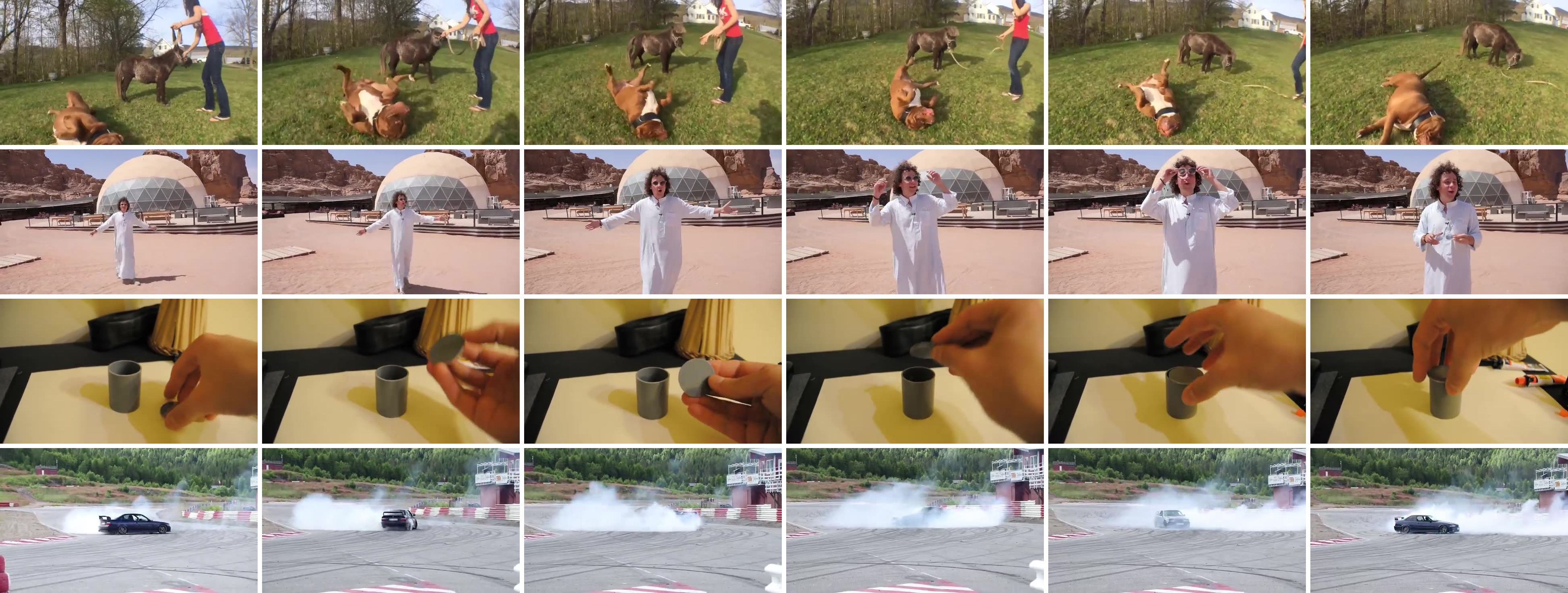}
    \caption{Examples of real videos collected from YouTube-VOS 2019~\cite{vos2019}.}
    \label{fig:reals_sup}
\end{figure*}

% \zliu{Full Deepfakes paragraph rewrote to align with version 2 facts, plz double check}
\paragraph{I. Full Deepfakes}
% =============== Ziyi rewrite ==================
The Full Deepfake category in \deepfakename includes two types: Text-to-Video (T2V) and Text-and-Image-to-Video (IT2V).

\textit{Text-to-Video (T2V).}
For T2V, we include a range of open-source generators, such as CogVideoX~\cite{yang2025cogvideoxtexttovideodiffusionmodels}, Hunyuan Video~\cite{kong2025hunyuanvideosystematicframeworklarge}, LTX-Video~\cite{hacohen2024ltxvideorealtimevideolatent}, Mochi~1 by Genmo~\cite{genmo2024mochi}, Open-Sora~\cite{zheng2024opensorademocratizingefficientvideo}, Wan2.1~\cite{wan2025wanopenadvancedlargescale}, and AllegroAI~\cite{zhou2024allegro}, as well as commercial generators Sora~\cite{openai2024sora} and Veo2~\cite{google2025veo2}, all released between 2024 and 2025.
To ensure diverse and semantically rich content, we generate prompts for all T2V methods using Google Gemini~\cite{geminiteam2025geminifamilyhighlycapable}.

\textit{Text-and-Image-to-Video (IT2V).}
For IT2V, we include models that support dual-modality conditioning (text and image), such as Open-Sora~\cite{zheng2024opensorademocratizingefficientvideo} and Wan2.1~\cite{wan2025wanopenadvancedlargescale}.
We first extract the initial frame from real videos (e.g., from Pexels~\cite{McCallumCorranPexelvideos}) and then query a VLM (Gemini) to produce a continuation prompt conditioned on this frame.
Both the prompts and the corresponding conditioning frames will be included in our metadata.

\begin{figure*}
    \centering
    \includegraphics[width=1.0\textwidth]{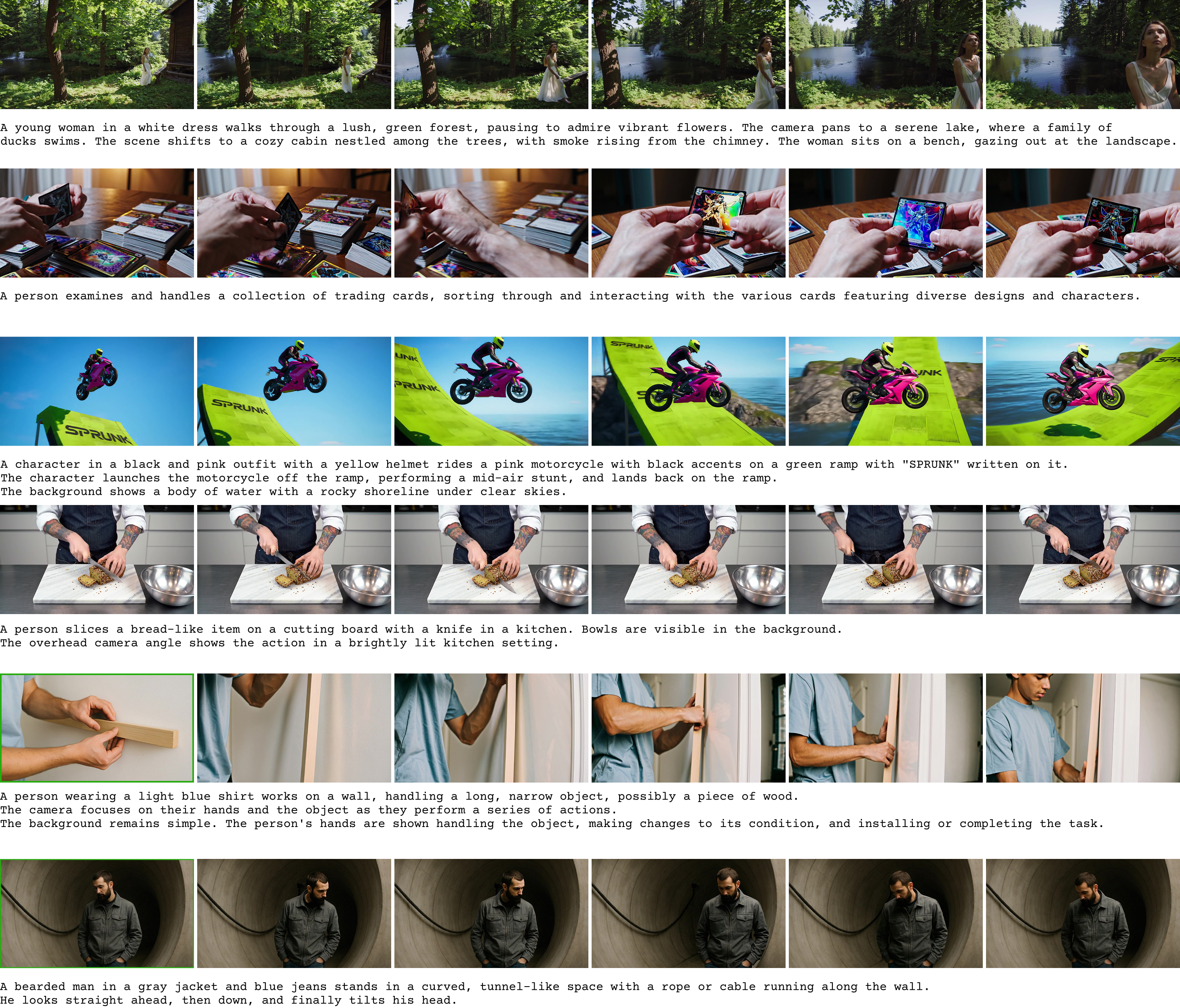}
    \caption{Examples of \textbf{Full Fakes}, including T2V (first two rows from SoRA~\cite{openai2024sora} and second two rows are from Veo2~\cite{google2025veo2}) and IT2V from SoRA~\cite{openai2024sora}. Green strokes indicate the condition image used in IT2V.}
    \label{fig:fullfakes_sup}
\end{figure*}

\paragraph{II. \deepfakename}

\paragraph{Spatial \deepfakename}

In the spatial subset of {\deepfakename{}}, we focus on methods that alter regional information within the video content.

\textit{FaceSwap Generation}.
To create FaceSwap deepfakes, we replaced the face in a target video with that from a source face image.
This was achieved using the InsightFace library~\cite{ren2023pbidr, guo2021sample, gecer2021ostec, an_2022_pfc_cvpr, an_2021_pfc_iccvw, deng2020subcenter, Deng2020CVPR, guo2018stacked, deng2018menpo, deng2018arcface}, a widely used tool for face detection and swapping.
Given the special requirement of the facial information in the video we draw our target videos from Celeb-DF dataset~\cite{li2020celebdflargescalechallengingdataset} and source ones from the CelebA dataset~\cite{liu2015faceattributes} and an additional set of celebrity face images collected in 2022.
To ensure gender consistency between the source and target frames, we prompt VLM with: \textit{"What is the gender of the person in the image?"} and filter the unmatched pairs.

\textit{Inpainting Generation.}
Inpainting-based manipulations simulate object removal within video scenes.
We include two sources of inpainted videos.

First, we generate new inpainted FakeParts using two recent video inpainting methods, DiffuEraser~\cite{li2025diffueraserdiffusionmodelvideo} and ProPainter~\cite{zhou2023propainter}:
\begin{enumerate}
    \item we extract the first frame from each video and query a vision-language model (VLM) with the prompt \textit{``What is one interesting object in the image that is neither too small to be noticeable nor so large that it occupies almost the entire frame?''} to identify a salient object for removal;
    \item we segment the selected object across all frames using Grounded-SAM-2~\cite{ravi2024sam2segmentimages}, obtaining a temporally consistent mask;
    \item we apply DiffuEraser or ProPainter to remove the object and fill the masked regions, producing realistic object-removal deepfakes.
\end{enumerate}

Second, we incorporate the language-driven inpainting dataset from ROVI~\cite{wu2024languagedrivenvideoinpaintingmultimodal}, where videos are inpainted using E2FGVI~\cite{wu2024languagedrivenvideoinpaintingmultimodal} under carefully curated prompts and parameter settings.
These ROVI inpainted clips are included as part of our inpainting FakeParts, complementing our own DiffuEraser/ProPainter-based generations with an additional, independently constructed inpainting pipeline.

\textit{Outpainting.} Unlike mainstream image inpainting and outpainting methods~\cite{lugmayr2022repaint,anciukevivcius2023renderdiffusion}, which primarily extrapolate content beyond image boundaries, video outpainting introduces additional complexities related to camera motion and 3D scene consistency.
Recent methods address this by explicitly integrating camera control into video generation~\cite{xu2024camco,he2024cameractrl,wang2024motionctrl,wang2024akira}.
In this section, we adopt the AkiRA model~\cite{wang2024akira}, built upon the SVD framework~\cite{blattmann2023stable}, which generates video content consistent with specified camera trajectories.
Specifically, we apply backwards camera tracking as conditioning input, enabling coherent extrapolation at frame boundaries.

\begin{figure*}
    \centering
    \includegraphics[width=1.0\textwidth]{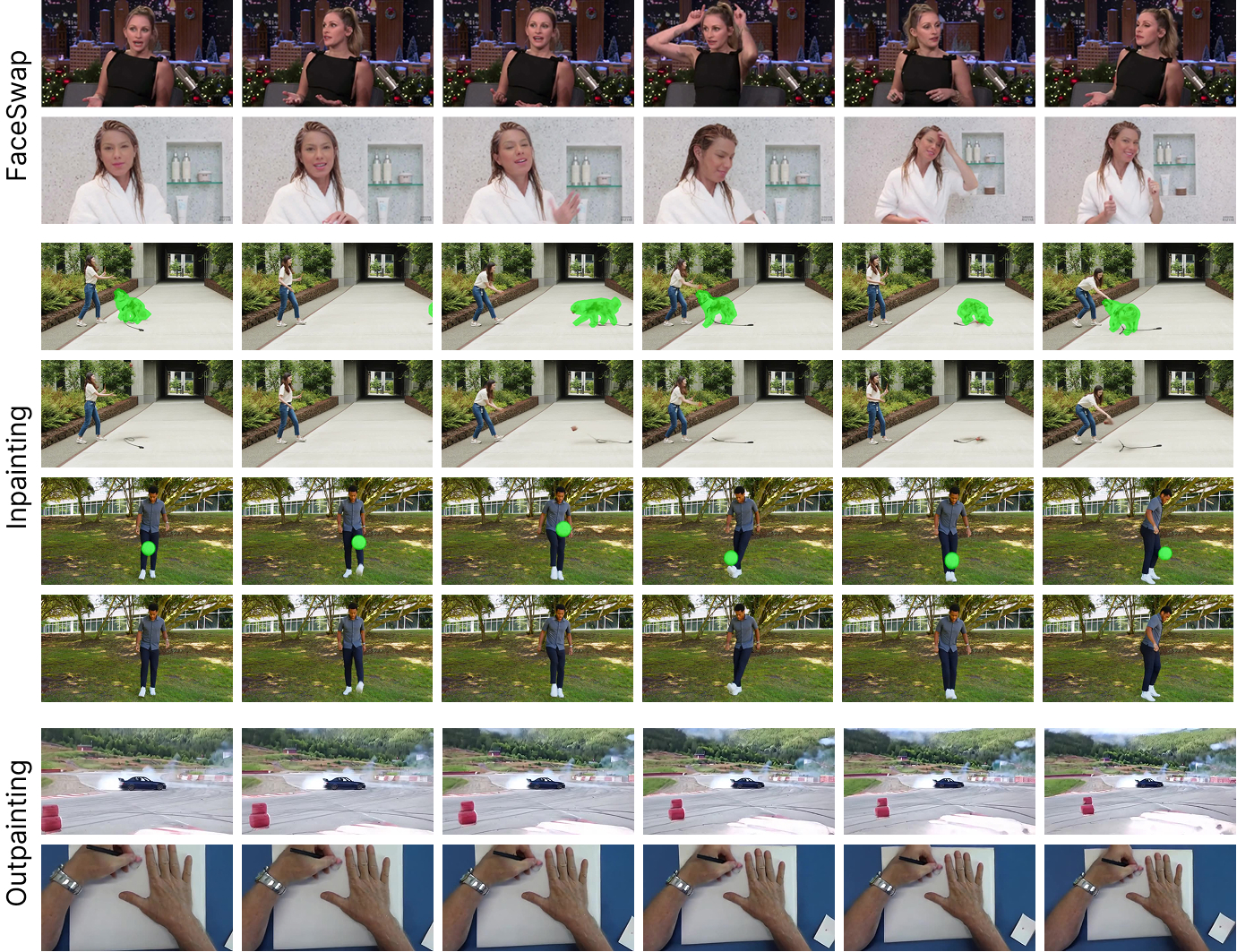}
    \caption{Examples of \textbf{Spatial \deepfakename{}}, including FaceSwap, inpainting, and outpainting methods. In inpainting, the green mask indicates the region to be removed.}
    \label{fig:spatial_fakeparts_sup_1}
\end{figure*}

\paragraph{Temporal \deepfakename}

In the temporal subset of {\deepfakename}, we focus on methods that complete or modify content along the time axis, such as frame interpolation.

\textit{Interpolation Generation.} For interpolation-based manipulations, we employ the Framer model~\cite{wang2024framerinteractiveframeinterpolation} to synthesize smooth transitions between non-consecutive frames, simulating temporal manipulations.
From each source video the first and last frames are provided to Framer, which generates 21 intermediate frames to fill the temporal gaps and produce a temporally manipulated video sequence.

\textit{Extrapolation Generation.}
To generate temporally extrapolated videos, we employ the Cosmos Predict Video2World framework~\cite{nvidia2025cosmospredict2github, nvidia2025cosmosworldfoundationmodel}, which predicts future motion and scene evolution.
Given the final 50 frames of a captured sequence, together with their caption as prompt, the model synthesises subsequent frames to extend each clip to 120 frames.
The resulting 25~FPS, 5-second videos preserve the original scene context while introducing plausible forward dynamics, forming our extrapolation subset.

\paragraph{Style \deepfakename}
The style subset of {\deepfakename} targets manipulations that retain the semantic content of the video but modify its appearance through style transfer techniques.

\textit{Style Change Generation.}
We use the diffusion-based RAVE model~\cite{kara2023raverandomizednoiseshuffling} to alter the colour and appearance of animals in videos from the Animal Kingdom dataset~\cite{Ng_2022_CVPR}.
For each video, we randomly sample a frame, caption it using PaLI-Gemma~2~\cite{steiner2024paligemma2familyversatile}, and prompt RAVE with: \textit{``Change the colour of the animal in the video.''}
In addition, we include object-specific colour edits generated with AnyV2V~\cite{ku2024anyv2v}.
In this case, frames are first captioned with VideoChat-R1~\cite{li2025videochatr1enhancingspatiotemporalperception}, from which we identify the object to edit and a target colour description.
We then construct prompts such as \textit{``Change the colour of the car to blue while keeping the rest of the scene unchanged.''}, ensuring that only local appearance is modified while preserving global scene semantics.

\begin{figure*}
    \centering
    \includegraphics[width=1.0\textwidth]{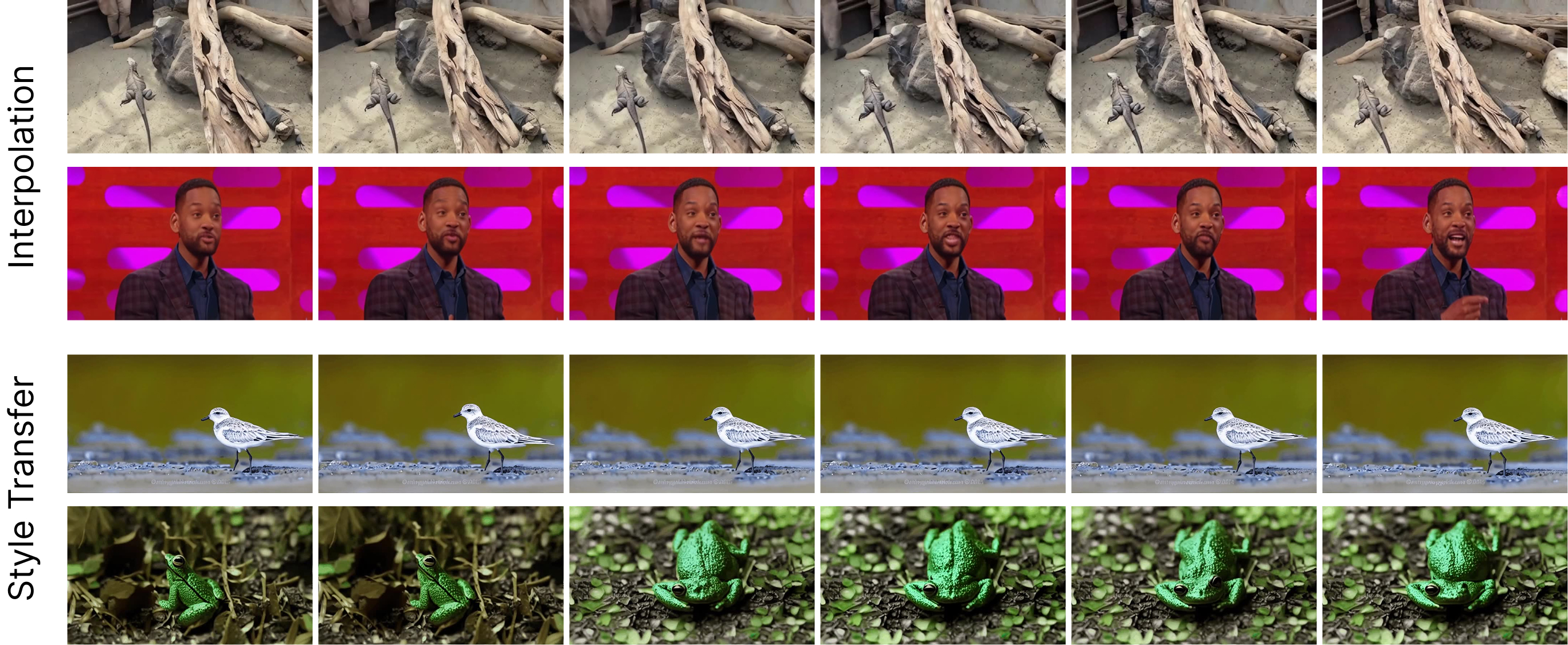}
    \caption{Examples of \textbf{Temporal and Style \deepfakename{}}, including Frame interpolation and Style Transfer.}
    \label{fig:spatial_fakeparts_sup_2}
\end{figure*}

\section{Metadata to be Released}

We will release all metadata necessary to reproduce each generation task in our study. 
All metadata will be uploaded together with the dataset to our public HuggingFace repository.

%----------------------------------------
\paragraph{Metadata for T2V Generation}

For the text-to-video (T2V) generation task, we provide the full set of textual prompts used as model inputs.
\begin{itemize}
    \item \textbf{Format:} CSV
    \item \textbf{Fields:}
    \begin{itemize}
        \item \texttt{id}
        \item \texttt{prompt\_text}
    \end{itemize}
\end{itemize}

%----------------------------------------
\paragraph{Metadata for TI2V Generation}

For the text-and-image-to-video (TI2V) generation task, we provide all multimodal prompts used in the experiments.
\begin{itemize}
    \item \textbf{Format:} CSV
    \item \textbf{Fields:}
    \begin{itemize}
        \item \texttt{id}
        \item \texttt{prompt\_text}
        \item \texttt{image\_reference}  \quad (reference to the corresponding input image)
    \end{itemize}
\end{itemize}

%----------------------------------------
\paragraph{Metadata for Real Videos Used in Extrapolation}

For real videos used as the basis for the Extrapolation task, we provide structural video clips’ metadata and aligned captions.
\begin{itemize}
    \item \textbf{Format:} CSV
    \item \textbf{Fields:}
    \begin{itemize}
        \item \texttt{name}
        \item \texttt{fps}
        \item \texttt{width}
        \item \texttt{height}
        \item \texttt{caption}
    \end{itemize}
\end{itemize}

%----------------------------------------
\paragraph{Metadata for Real Videos Used in Style Change}

For real videos used as style sources in the Style Change task, we release their corresponding metadata,  captions and object concerned.
\begin{itemize}
    \item \textbf{Format:} CSV
    \item \textbf{Fields:}
    \begin{itemize}
        \item \texttt{name}
        \item \texttt{fps}
        \item \texttt{width}
        \item \texttt{height}
        \item \texttt{caption}
        \item \texttt{objects}
    \end{itemize}
\end{itemize}

%----------------------------------------
\paragraph{Metadata for Extrapolation and Style-Change Prompts}

We additionally release the prompt format and pipelines used specifically for the Extrapolation and Style-Change generation tasks.
\begin{itemize}
    \item \textbf{Format:} Jupyter Notebook
\end{itemize}

%----------------------------------------
\paragraph{Release Format and Location}

All metadata will be released in CSV format and hosted within the same HuggingFace dataset repository as the main data, ensuring full reproducibility.

\end{document}